\documentclass{article} 
\usepackage{colm}
\usepackage[colorlinks = true,
            linkcolor = blue,
            urlcolor  = blue,
            citecolor = blue,
            anchorcolor = blue]{hyperref}
\usepackage[utf8]{inputenc} 
\usepackage[T1]{fontenc}    
\usepackage{url}            
\usepackage{booktabs}       
\usepackage{amsfonts}       
\usepackage{nicefrac}       
\usepackage{microtype}      
\usepackage{amsmath}
\usepackage{amssymb}
\usepackage{multirow}
\usepackage{tabularx}
\usepackage{makecell}
\usepackage{array}
\newcolumntype{C}[1]{>{\centering\arraybackslash}m{#1}}
\usepackage{xcolor}
\usepackage{colortbl}
\usepackage{graphicx}
\usepackage{textcomp}       
\usepackage{pifont}
\usepackage[most]{tcolorbox}
\usepackage{listings}
\usepackage{fancyvrb}
\usepackage{fvextra}
\usepackage{caption}
\usepackage{adjustbox}
\usepackage{wrapfig}
\usepackage{float}
\usepackage{subfig}
\usepackage{framed}
\usepackage{xspace}
\usepackage[shortlabels]{enumitem}

\newcommand\blfootnote[1]{%
  \begingroup
  \renewcommand\thefootnote{}\footnote{#1}%
  \addtocounter{footnote}{-1}%
  \endgroup
}

\title{ARISE-RL: Agentic Rubric-Grounded Iterative Self-Evolution with Reinforcement Learning}

\author{Fanrui Zhang$^{1}$, Ruixue Ding$^{1\ast}$, Qiang Zhang$^{1}$, Xi Chen$^{1}$, Boli Chen$^{1}$, Shihang Wang$^{1}$,\\
\textbf{Hongmin Zhan, Jinxin Bian$^{2}$, Xingchao Li$^{2}$, Peijin Zheng$^{2}$, Hao Cheng$^{2}$, Pengjun Xie$^{1}$,}\\
\textbf{Kaipeng Zhang, Jiawei Liu, Zheng-Jun Zha}\\
\\
$^{1}$\includegraphics[height=0.4cm]{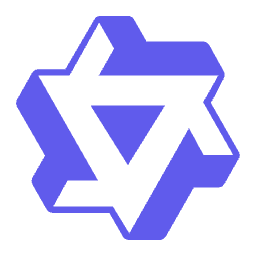} Alibaba ATH Token Foundry \quad $^{2}$Hema, Alibaba Group
}

\colmfinalcopy 
\begin{document}

\maketitle

\blfootnote{$^{\ast}$ Project leader}


\begin{abstract}

Training open-ended agents via reinforcement learning (RL) is hindered by the lack of verifiable gold answers and scalable rubrics. Moreover, even near the model’s capability boundary, long-horizon open-ended agentic tasks often yield brittle and unstable rewards, resulting in weak or noisy rollout contrast that obscures fine-grained optimization signals for group-based policy learning. To address these challenges, we propose ARISE-RL, a novel full-cycle self-evolution framework that couples a task/rubric \textit{Generator} and a reasoning \textit{Solver} through rubric-mediated co-evolution. The Generator grounds tool-related rubric criteria in real tool observations and is rewarded for producing valid, intermediate-difficulty tasks aligned with the Solver's evolving capability boundary. The Solver, in turn, learns from fine-grained rubric satisfaction signals through multi-step reasoning and tool use. We further introduce \textit{Reward-Gated Self-Evolution Distillation} (RG-SED), which selectively distills a memory-augmented variant of the same policy back into itself only when the memory yields empirical reward improvement, thereby reducing distribution mismatch and avoiding blind imitation of noisy guidance. Finally, to support rigorous evaluation, we present ECR-Bench, an expert-calibrated rubric benchmark suite covering single-tool deep research and multi-tool travel planning. Extensive experiments demonstrate that ARISE-RL consistently achieves robust and stable overall state-of-the-art performance across all evaluated benchmarks.
The code is available at \href{https://github.com/Alibaba-NLP/qqr}{https://github.com/Alibaba-NLP/qqr}.
\end{abstract}

\begin{figure}[!t]
  \centering
  \includegraphics[width=0.8\linewidth]{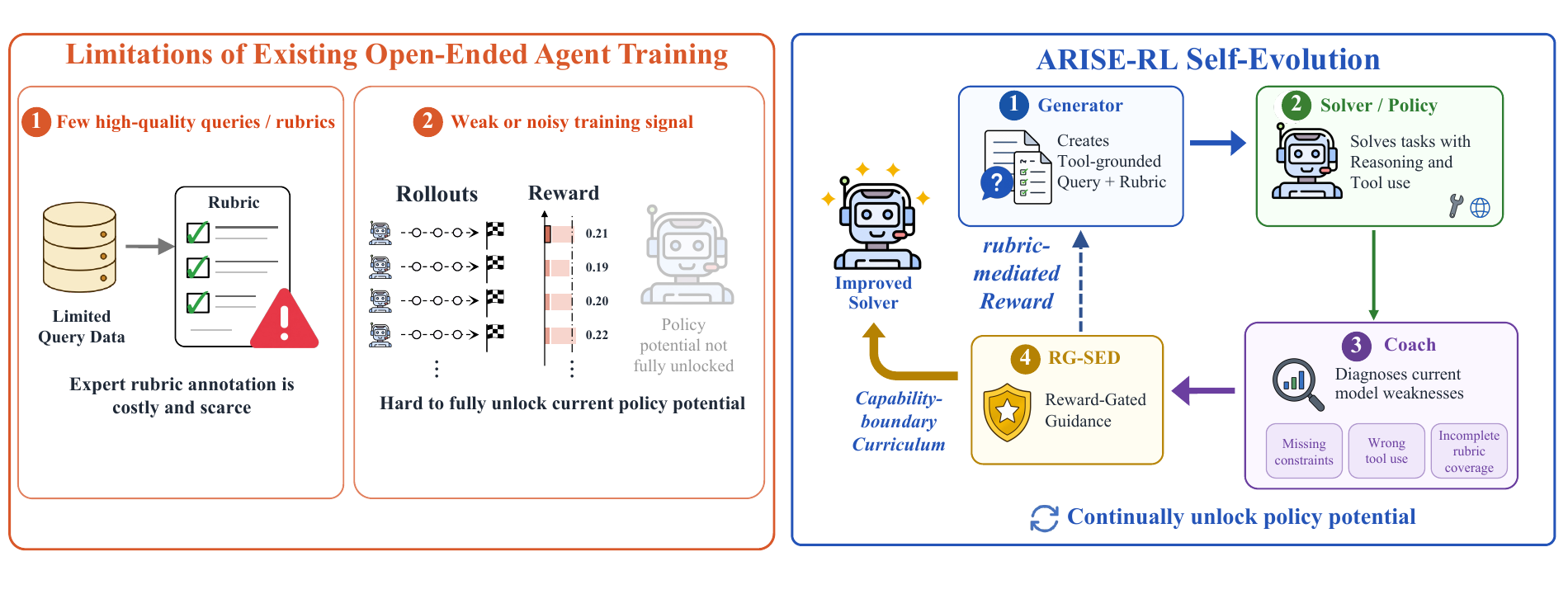}
  \caption{Motivation of ARISE-RL. Existing long-horizon open-ended agent training is constrained by scarce high-quality queries and expert rubrics, as well as brittle and noisy rewards that provide weak rollout contrast and limit policy improvement. ARISE-RL introduces a rubric-mediated self-evolution loop, where the Generator produces tool-grounded queries and rubrics near the Solver’s capability boundary, while the Solver improves through fine-grained rubric rewards and reward-gated guidance. This design reduces reliance on human supervision and expands capabilities.}
  \vspace{-3mm}
  \label{fig:teaser}
\end{figure}

\section{Introduction}


Large language models (LLMs) are rapidly evolving into autonomous agents capable of planning, tool use, and executing complex real-world tasks~\citep{yao2022react,qin2024toolllm,zhang2026arenarl}. While reinforcement learning (RL) has successfully optimized agents in verifiable domains like mathematics and code generation via exact automated feedback~\citep{shao2024deepseekmath,jimenez2024swebench}, many valuable real-world tasks (e.g., deep research, travel planning) are inherently open-ended. 
These tasks lack a single gold standard and require satisfying multi-dimensional, interacting criteria such as factual grounding, constraint satisfaction, reasoning coherence, and utility, making their optimization exceptionally challenging.

This open-ended nature exposes two fundamental bottlenecks in current agentic RL paradigms. First, a \textit{data--evaluation bottleneck} arises because high-quality, open-ended training data and expert rubrics are expensive to scale. Existing benchmarks primarily target deterministic environments with verifiable end states or final outputs~\citep{zhou2024webarena}. Recent open-ended suites~\citep{he2025vitabench,du2025deepresearchbench} broaden the evaluation scope, but largely remain static testbeds for post-hoc assessment, lacking the training data required for continuous, fine-grained optimization.

Second, an \textit{optimization bottleneck} persists even when queries are placed near the model's capability boundary. Open-ended agentic tasks typically involve long-horizon, multi-turn tool use and interaction with external environments, where reward signals are often brittle and unstable. Consequently, vanilla RL may still suffer from \textit{advantage collapse}: sampled rollouts provide only weak or noisy contrast, making it difficult to derive fine-grained optimization signals that indicate which reasoning behaviors and tool-use decisions should be reinforced. 
This confines optimization to local refinement within the model's existing capability region, rather than enabling sustained expansion of its capability frontier.
A natural remedy is to introduce a stronger teacher who provides demonstrations or guidance. 
However, conventional off-policy distillation is limited in multi-turn agentic settings: the distribution gap between a fixed teacher and an evolving learner can accumulate through multi-step reasoning and tool use, leading to unstable training~\citep{ye2026policy}. 
Moreover, teacher guidance in open-ended tasks is not always reliable, and blind imitation may transfer incorrect tool choices, missed constraints, or low-quality reasoning patterns into the student's policy. 
These limitations call for a closed-loop paradigm that can generate tasks, calibrate difficulty, and verify guidance usefulness before incorporating it into policy learning.

To this end, we propose \textbf{ARISE-RL} (\textbf{A}gentic \textbf{R}ubric-Grounded \textbf{I}terative \textbf{S}elf-\textbf{E}volution with \textbf{R}einforcement \textbf{L}earning), a full-cycle self-evolution RL framework for open-ended agents illustrated in Figure~\ref{fig:teaser}. ARISE-RL replaces reliance on static external data with a closed loop where task generation, task solving, and rubric-based evaluation co-evolve. It consists of a \textit{Generator}, which produces open-ended queries and judging rubrics, and a \textit{Solver}, which learns to solve these queries through multi-step reasoning and tool use under rubric-based rewards.
Rubrics serve as the interface between task construction and policy optimization. As the Solver improves, the Generator is driven to create tasks that better match the Solver's evolving capability frontier, forming a rubric-mediated Generator--Solver co-evolution process. To ensure reliable self-generated supervision, ARISE-RL introduces \textit{tool-grounded rubric construction}: the Generator must invoke relevant tools before writing criteria that depend on tool outputs. This prevents unsupported or hallucinated tool-related rubric items and aligns generated supervision with the actual tool environment.
ARISE-RL further calibrates task hardness with a difficulty-shaped reward. 
For each generated query, multiple Solver attempts estimate empirical solvability, and the Generator is rewarded most when success rates are intermediate rather than uniformly high or low. 
This encourages tasks near the Solver's capability boundary, where samples are both challenging and informative for reinforcement learning.


\begin{figure}[!t]
  \centering
  \includegraphics[width=\textwidth]{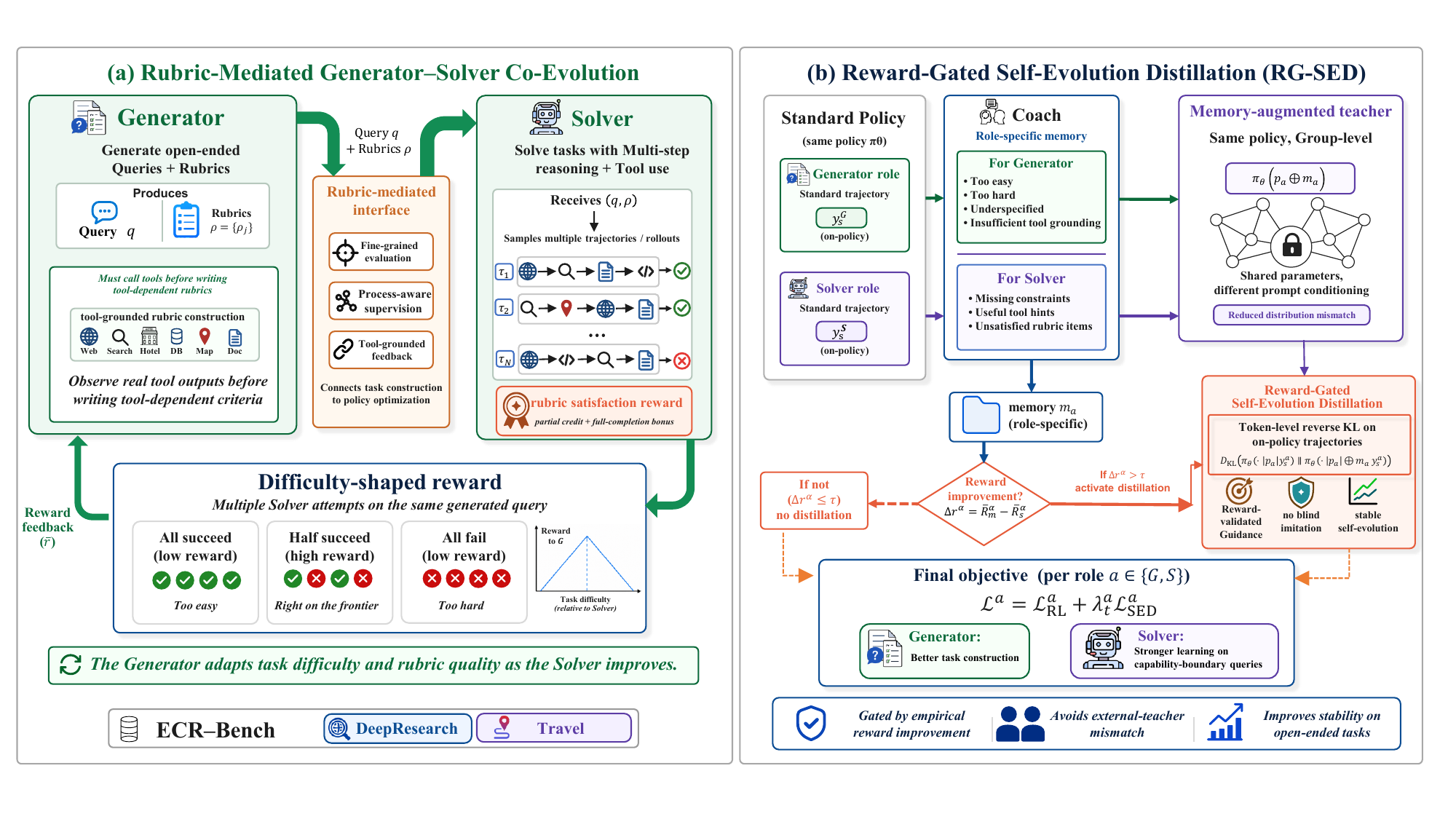}
   \caption{
    Overview of ARISE-RL. At the \emph{global} level, a Generator creates open-ended queries and tool-grounded rubrics, while a Solver learns to satisfy them through reasoning and tool use. The rubric couples task generation with policy optimization, and a difficulty-shaped reward keeps tasks near the Solver's capability frontier. At the \emph{local} level, RG-SED builds a transient teacher from the same policy with group-level coach memory, and distills its behavior only when memory-augmented rollouts empirically improve reward.
    }
  \label{framework}
  \vspace{-3mm}
\end{figure}

At the local optimization level, we propose {Reward-Gated Self-Evolution Distillation} (RG-SED), a group-level self-distillation mechanism that continually unlocks policy potential. For the Generator, RG-SED distills reward-improving task-construction patterns, enabling generated queries and rubrics to align with the Solver's capability boundary more rapidly. For the Solver, RG-SED strengthens learning on capability-boundary queries, mitigating weak or low-variance updates that may arise in standard group-based RL on highly informative samples. Unlike methods that rely on a fixed external teacher, RG-SED constructs a transient enhanced teacher from the same policy under memory-augmented conditioning, and activates distillation only when its rollouts outperform standard rollouts. Once activated, RG-SED applies token-level reverse KL on on-policy trajectories, allowing the policy to internalize reward-validated high-confidence behaviors. This reward-gated design avoids distribution mismatch and prevents blind imitation of noisy guidance.

Beyond the training framework, we construct ECR-Bench, an {Expert-Calibrated Rubric Benchmark} suite for open-ended agent evaluation. It contains {ECR-DeepResearch}, with 100 expert-calibrated single-tool research queries, and {ECR-Travel}, a multi-tool travel-planning benchmark covering five task types: route planning, transportation comparison, nearby POI search, one-day itinerary planning, and multi-day itinerary planning. ECR-Bench evaluates both final response quality and process-level tool-use correctness.
We evaluate ARISE-RL on single-tool deep research and multi-tool task planning. The former is measured by rubric score rate, while the latter is evaluated by task pass rate. Across ECR-DeepResearch, ECR-Travel, and existing open-ended benchmarks such as ResearchRubrics~\citep{sharma2025researchrubrics} and VitaBench~\citep{he2025vitabench}, ARISE-RL achieves the best average performance and consistent robust gains on 8B/9B-scale models.

Our main contributions are summarized as follows. 
(1) We propose ARISE-RL, a full-cycle self-evolution RL framework that unifies task generation, rubric construction, task solving, and reinforcement learning into a closed loop, enabling open-ended agents to continually expand their capability boundaries with reduced reliance on large-scale human-authored data. 
(2) We introduce {Reward-Gated Self-Evolution Distillation} (RG-SED), which constructs a transient enhanced teacher from the same policy under group-level coach memory and activates distillation only when the memory empirically improves reward, mitigating advantage collapse, distribution mismatch, and negative transfer from noisy guidance. 
(3) We construct {ECR-Bench}, an Expert-Calibrated Rubric Benchmark suite covering single-tool deep research and multi-tool travel planning, providing fine-grained, process-aware, and expert-calibrated evaluation criteria for open-ended agents.

\section{Related Work}
\label{sec:related}

\paragraph{Self-Evolving Agents.}
Recently, self-evolution has emerged as a paradigm for autonomous LLM improvement~\citep{fang2025selfevolvesurvey,yue2026dr,liu2025agent0,huang2025r,lu2025search}. EvolveR~\citep{wu2025evolver} couples offline self-distillation with online interaction over a repository of strategic principles, while ASL~\citep{sun2025towards} unifies prompt generation, policy learning, and generative reward modeling for search agents. RAGEN~\citep{wang2025ragen} studies self-evolution in multi-turn RL. While prior frameworks demonstrate the potential of self-bootstrapping, they mainly focus on closed-form, single-tool, or reasoning-centric settings, and thus fall short of open-ended agentic tasks that require reliable tool grounding, adaptive task generation, and fine-grained rubric-based evaluation. 
ARISE-RL is the first self-evolution framework tailored to such tasks, unifying task generation, rubric construction, and policy learning within a closed loop.

\paragraph{Off-Policy Distillation and Self-Distillation.}
Distillation from a stronger fixed teacher~\citep{hinton2015distilling,gou2021knowledge,lu2025search} is a common way to bootstrap difficult RL settings. In multi-turn agentic training, however, the importance-sampling correction between a static teacher and the on-policy student compounds across tool-use turns and quickly destabilizes optimization, a failure mode reported in OPCD~\citep{ye2026policy}. A fixed teacher additionally imposes a static ceiling on the student and provides no mechanism for rejecting noisy guidance. Self-distillation~\citep{zhang2019your,allen2020towards} sidesteps the cross-model gap but, in its standard form, lacks a principled criterion for \emph{when} to distill. 
RG-SED constructs the teacher from the same policy under a query-specific coach-augmented prompt, naturally bounding the student--teacher distribution gap. Distillation is activated only when coach-augmented rollouts improve empirical reward, and useful guidance is internalized via token-level reverse KL on on-policy trajectories to preserve training--deployment consistency.


\section{Method}
\label{sec:method}

As shown in Figure~\ref{framework}, \textbf{ARISE-RL} trains open-ended agents through a self-evolving loop of task generation, rubric construction, and policy learning. Globally, a \textit{Generator} creates open-ended queries and rubrics near the \textit{Solver}'s capability boundary, while the Solver learns through reasoning and tool use. Locally, \textit{Reward-Gated Self-Evolution Distillation} (RG-SED) selectively distills reward-improving behaviors for both roles, yielding more informative tasks and stronger open-ended agents.

\subsection{Rubric-Mediated Generator--Solver Co-Evolution}

ARISE-RL builds a dynamic rubric-mediated co-evolution loop between a task Generator and a Solver. 
At each iteration, the Generator produces an open-ended query $q$ and a set of evaluation rubrics $\rho=\{\rho_j\}_{j=1}^{M}$, while the Solver attempts to answer the query through multi-step reasoning and tool use. 
The generated rubric serves as the coupling interface between task construction and policy optimization: it provides explicit fine-grained evaluation criteria, defines process-aware supervision for Solver learning, and offers tool-grounded feedback for Generator training.

\paragraph{Generator reward.}
The Generator is encouraged to produce tasks that are tool-grounded, format-valid, and appropriately challenging. Given a Generator trajectory $\xi$, we define its reward as:
\begin{equation}
    R_{\mathrm{G}}(q,\rho,\xi)
    = R_{\mathrm{tool}}(\xi) R_{\mathrm{fmt}}(q,\rho)
    (1+ R_{\mathrm{diff}}(q,\rho)).
\end{equation}
where $R_{\mathrm{tool}}$ and $R_{\mathrm{fmt}}$ are binary gates. The tool-grounding gate requires the Generator to observe at least one real tool output before specifying tool-dependent rubric criteria:
Let $\mathcal{T}(\xi)$ denote the event that trajectory $\xi$ contains the corresponding tool observation. We define
\begin{equation}
    R_{\mathrm{tool}}(\xi)
    =
    \mathbb{I}\!\left[\mathcal{T}(\xi)\right].
\end{equation}
This prevents the Generator from hallucinating tool-related requirements that are unsupported by actual tool evidence. The format gate ensures that the generated task can be parsed and executed by the downstream Solver:
\begin{equation}
    R_{\mathrm{fmt}}(q,\rho)
    =
    \mathbb{I}
    \big[
    q \neq \varnothing
    \;\wedge\;
    \rho \text{ is a non-empty list}
    \big].
\end{equation}
Once both gates are satisfied, the Generator receives a difficulty-shaped reward. For each generated query, we independently sample the Solver $K$ times. Let $r_i^{\mathrm{S}}$ denote the Solver reward for the $i$-th rollout, and let $\gamma$ be the success threshold. The number of successful rollouts is defined as
\begin{equation}
    c
    =
    \sum_{i=1}^{K}
    \mathbb{I}
    \left(
    r_i^{\mathrm{S}} \ge \gamma
    \right).
\end{equation}
We then define the difficulty reward as
\begin{equation}
    R_{\mathrm{diff}}(q,\rho)
    =
    2 \cdot
    \max
    \left(
    0,\,
    1 -
    \frac{|c-K/2|}{K/2}
    \right).
\end{equation}
This reward is maximized when approximately half of the Solver rollouts succeed, and decreases to zero when all rollouts either succeed or fail. It therefore drives the Generator toward intermediate-difficulty tasks near the Solver's current capability boundary, where group-based policy optimization receives the most informative reward variation.

\paragraph{Solver reward.}
Given a Solver trajectory $\tau_i$ and a rubric set $\rho=\{\rho_j\}_{j=1}^{M}$, we assess whether each rubric item is satisfied using an LLM judge. Specifically, we define
\begin{equation}
    z_{ij}
    =
    \mathbb{I}
    \left[
    \tau_i \text{ satisfies } \rho_j
    \right],
\end{equation}
and compute the rubric satisfaction rate as
\begin{equation}
    s_i
    =
    \frac{1}{M}
    \sum_{j=1}^{M}
    z_{ij}.
\end{equation}
The rubric-mediated reward combines partial-credit supervision with a full-completion bonus:
\begin{equation}
    r_i^{\mathrm{S}} = \alpha s_i + (1-\alpha)\mathbb{I}[s_i=1],
    \qquad r_i^{\mathrm{S}} \in [0,1].
\end{equation}
where $\alpha$ balances partial rubric satisfaction and the full-completion bonus. This reward provides fine-grained supervision while retaining an explicit incentive for satisfying all criteria. It is used for group-based Solver optimization and reused as the empirical difficulty signal for Generator training.

\paragraph{Co-evolution dynamics.}
The Generator and Solver rewards form an adaptive self-evolving curriculum. The Solver learns to satisfy fine-grained rubrics, while the Generator is driven to produce valid, tool-grounded tasks near the Solver's capability boundary, yielding a closed-loop optimization process for continual open-ended agent learning.

\subsection{Role-Conditioned Reward-Gated Self-Evolution Distillation}

Although rubric-mediated co-evolution shapes a progressively adaptive curriculum, both the Generator and the Solver can still encounter local optimization inefficiencies. For the Generator, reward feedback from Solver performance may be sparse or delayed, making it difficult to quickly identify task construction patterns that match the Solver's current capability boundary. For the Solver, capability-boundary long-horizon tasks are highly informative but may yield weak or low-variance group-based learning signals, limiting the efficiency of policy improvement. To address these issues, we introduce RG-SED, a role-conditioned group-level self-distillation mechanism applied to both roles.

Let $a \in \{\mathrm{G}, \mathrm{S}\}$ denote the role, corresponding to the Generator or the Solver. For each role, the current deployment policy first samples trajectories under the original role-specific prompt $p_a$:
\begin{equation}
    y_s^{a} \sim \pi_{\theta}^{a}(\cdot \mid p_a),
\end{equation}
where $y_s^{a}$ denotes a standard trajectory, instantiated as a Generator trajectory for $a=\mathrm{G}$ and a Solver trajectory for $a=\mathrm{S}$.
A coach then analyzes the feedback associated with these trajectories and produces role-specific memory $m_a$. For the Generator, this memory summarizes task-construction feedback, such as whether the generated query was too easy, too difficult, underspecified, or insufficiently grounded in tool observations. Such memory helps the Generator adjust future queries and rubrics toward the Solver's current capability frontier. For the Solver, the memory summarizes query-specific learning signals, such as missing constraints, useful tool hints, or rubric items that should be explicitly addressed. Such memory strengthens learning on capability-boundary queries without introducing an external teacher.
Conditioning the same policy on the role-specific memory yields a transient memory-augmented teacher:
\begin{equation}
    y_m^{a} \sim \pi_{\theta}^{a}(\cdot \mid p_a \oplus m_a).
\end{equation}
Since the student and teacher share the same parameters and differ only in prompt conditioning, their distributional gap is naturally constrained. This avoids the severe mismatch of conventional off-policy distillation while allowing the policy to benefit from reward-improving self-generated guidance.

RG-SED activates distillation only when the group-level memory-augmented trajectories provide empirical reward improvement. For each role $a$, we compute the reward gap:
\begin{equation}
    \Delta_r^{a}
    =
    \bar{R}_{m}^{a}
    -
    \bar{R}_{s}^{a},
\end{equation}
where $\bar{R}_{m}^{a}$ and $\bar{R}_{s}^{a}$ denote the average rewards of memory-augmented and standard trajectories, respectively. The distillation strength is controlled by a reward-gated coefficient:
\begin{equation}
    \lambda_t^{a}
    =
    \lambda_0^{a}
    \cdot
    \mathbb{I}(\Delta_r^{a} > \tau_a)
    \cdot
    w_a(t)
    \cdot
    d_a(t),
\end{equation}
where $\tau_a$ is the role-specific activation threshold, $w_a(t)$ is a warm-up schedule, and $d_a(t)$ is a decay schedule.
When activated, RG-SED applies token-level reverse KL regularization on the standard on-policy trajectories:
\begin{equation}
\begin{aligned}
\mathcal{L}_{\mathrm{SED}}^{a} ={} & \mathbb{E}_{y_s^{a} \sim \pi_{\theta}^{a}(\cdot \mid p_a)} \Bigg[ \sum_{t} D_{\mathrm{KL}} \Biggl( \pi_{\theta}^{a}(\cdot \mid h_t, p_a) \;\Bigg\|\; \\
& \mathrm{sg} \!\left[ \pi_{\theta}^{a}(\cdot \mid h_t, p_a \oplus m_a) \right] \Biggr) \Bigg],
\end{aligned}
\end{equation}
where $h_t$ denotes the trajectory history at step $t$, and $\mathrm{sg}[\cdot]$ stops gradients through the memory-augmented distribution. The final role-specific optimization objective is:
\begin{equation}
    \mathcal{L}^{a}
    =
    \mathcal{L}_{\mathrm{RL}}^{a}
    +
    \lambda_t^{a}
    \mathcal{L}_{\mathrm{SED}}^{a},
    \qquad
    a \in \{\mathrm{G}, \mathrm{S}\}.
\end{equation}

By gating distillation with empirical reward improvement, RG-SED avoids blind imitation of noisy guidance, mitigates distribution mismatch, and improves the stability and efficiency of open-ended agent self-evolution.

\section{ECR-Bench}
\label{sec:ecr-bench}

To evaluate open-ended agents with reliable and fine-grained criteria, we introduce \textbf{ECR-Bench}, an \textbf{E}xpert-\textbf{C}alibrated \textbf{R}ubric \textbf{Bench}mark suite. It contains two complementary domains: {ECR-DeepResearch} for single-tool open-ended research and {ECR-Travel} for multi-tool travel planning.







\paragraph{Stage I: Benchmark design and expert rubric annotation.}

We manually review all test queries and rubrics to ensure query clarity, appropriate difficulty, rubric specificity, coverage, and consistency between each query and its corresponding rubric. Queries or rubrics that fail the review are iteratively revised or removed to ensure evaluation reliability.

(1) ECR-DeepResearch evaluates web-search-based deep research, requiring agents to retrieve information, synthesize evidence, and produce structured research-style responses. It contains 100 high-quality test queries, each paired with an expert-calibrated rubric assessing factual grounding, evidence coverage, reasoning quality, completeness, and report structure. We use the rubric score ratio as the main metric.
(2) ECR-Travel evaluates multi-tool travel planning under realistic constraints, including time windows, budgets, user preferences, transportation feasibility, and weather conditions. It covers five task categories, with 100 test queries per category: route planning with multiple specified waypoints (\textit{Direction}); transportation-mode comparison (\textit{Compare}); nearby point-of-interest (POI) search (\textit{Search}); one-day trip planning in a single city (\textit{1-Day}); and multi-day trip planning (\textit{M-Day}). The tool set includes POI search, nearby search, navigation, web search, flight search, train-ticket search, and weather lookup. Each query is paired with an expert-calibrated rubric assessing both itinerary quality and tool-use correctness, with task success rate as the primary evaluation metric.
Together, ECR-DeepResearch and ECR-Travel evaluate complementary capabilities: evidence-grounded long-form research and constraint-aware multi-tool planning. ECR-Bench therefore provides an open-ended agent evaluation platform beyond fixed-answer benchmarking.

\paragraph{Stage II: Quality control.}
We apply a rule-augmented LLM quality checker to filter trajectories with formatting errors or logical inconsistencies. The checker strictly validates tool-call validity, dialogue correctness, and final-answer consistency. Failed samples are iteratively rewritten or removed, and all test queries and rubrics are manually reviewed to ensure reliability.

\begin{table}[!t]
\centering
\caption{Main results on four agentic benchmarks. We report rubric score rate on ResearchRubrics (RR) and ECR-DeepResearch (ECR-DR), and task pass rate on VitaBench and ECR-Travel. Avg. denotes the average across the four benchmarks, with VitaBench and ECR-Travel aggregated by averaging their corresponding sub-task scores.} 
\label{tab:main_results}
{\renewcommand{\arraystretch}{0.95}%
\resizebox{1.0\textwidth}{!}{
\begin{tabular}{l|c|c|cccc|ccccc|c}
\toprule
\multirow{2}{*}{\textbf{Method}}
& \multirow{2}{*}{\textbf{RR}}
& \multirow{2}{*}{\textbf{ECR-DR}}
& \multicolumn{4}{c|}{\textbf{VitaBench}}
& \multicolumn{5}{c|}{\textbf{ECR-Travel}}
& \multirow{2}{*}{\textbf{Avg.}} \\
\cmidrule(lr){4-7} \cmidrule(lr){8-12}
& &
& \textbf{Deliv.} & \textbf{In-Store} & \textbf{OTA} & \textbf{Cross}
& \textbf{Direction} & \textbf{Compare} & \textbf{Search} & \textbf{1-Day} & \textbf{M-Day}
& \\
\midrule
\multicolumn{13}{c}{\emph{Closed-source LLMs}} \\
\midrule
Gemini3-Pro~\citep{google2025gemini3}                & 0.473 & 0.687 & 0.410 & 0.443 & 0.380 & 0.246 & 0.187 & 0.642 & 0.738 & 0.290 & 0.378 & 0.494 \\
GPT-5~\citep{openai2025gpt5}                         & 0.316 & 0.543 & 0.498 & 0.504 & 0.327 & 0.210 & 0.100 & 0.565 & 0.758 & 0.317 & 0.410 & 0.418 \\
GPT-5.2~\citep{openai2025gpt52}                       & 0.434 & 0.713 & 0.342 & 0.273 & 0.295 & 0.071 & 0.027 & 0.567 & 0.609 & 0.276 & 0.398 & 0.442 \\
Claude-4.5-Sonnet~\citep{anthropic2025claude45}      & 0.427 & 0.541 & 0.473 & 0.467 & 0.419 & 0.287 & 0.398 & 0.783 & 0.713 & 0.243 & 0.260 & 0.465 \\
Claude-4.6-Sonnet~\citep{anthropic2025claude46}      & 0.467 & 0.623 & \textbf{0.598} & 0.472 & \textbf{0.426} & 0.273 & 0.223 & 0.683 & 0.710 & 0.236 & 0.393 & 0.495 \\
\midrule
\multicolumn{13}{c}{\emph{Open-source LLMs}} \\
\midrule
Qwen3-32B~\citep{yang2025qwen3}                      & 0.372 & 0.590 & 0.380 & 0.318 & 0.247 & 0.173 & 0.389 & 0.586 & 0.612 & 0.247 & 0.298 & 0.417 \\
Qwen3-235B~\citep{yang2025qwen3}                     & 0.421 & 0.652 & 0.450 & 0.398 & 0.318 & 0.216 & 0.452 & 0.658 & 0.687 & 0.273 & 0.340 & 0.475 \\
Qwen3.5-397B~\citep{qwen2026qwen35}                   & 0.418 & 0.628 & 0.580 & 0.512 & 0.358 & \textbf{0.293} & 0.512 & 0.756 & 0.748 & 0.319 & 0.410 & 0.508 \\
\midrule
\multicolumn{13}{c}{\emph{Self-Evolving Frameworks}} \\
\midrule
Qwen3-8B~\citep{yang2025qwen3}              & 0.236 & 0.412 & 0.143 & 0.117 & 0.026 & 0.013 & 0.387 & 0.346 & 0.490 & 0.213 & 0.282 & 0.267 \\
+Dr. Zero~\citep{yue2026dr}                          & 0.292 & 0.462 & 0.172 & 0.142 & 0.058 & 0.030 & 0.442 & 0.410 & 0.557 & 0.240 & 0.318 & 0.312 \\
+Absolute Zero~\citep{zhao2026absolute}              & 0.307 & 0.417 & 0.176 & 0.144 & 0.018 & 0.016 & 0.361 & 0.450 & 0.513 & 0.257 & 0.307 & 0.298 \\
\rowcolor{gray!20} +ARISE-RL                         & 0.343 & 0.513 & 0.202 & 0.167 & 0.094 & 0.055 & 0.498 & 0.483 & 0.617 & 0.273 & 0.357 & 0.358 \\
\addlinespace[2pt]
Qwen3.5-9B~\citep{qwen2026qwen35}            & 0.447 & 0.735 & 0.358 & 0.483 & 0.118 & 0.063 & 0.438 & 0.657 & 0.672 & 0.263 & 0.321 & 0.477 \\
+Dr. Zero~\citep{yue2026dr}                          & 0.464 & 0.761 & 0.454 & 0.578 & 0.272 & 0.103 & 0.524 & 0.747 & 0.744 & 0.312 & 0.403 & 0.531 \\
+Absolute Zero~\citep{zhao2026absolute}              & 0.452 & 0.748 & 0.406 & 0.504 & 0.308 & 0.080 & 0.452 & 0.676 & 0.667 & 0.323 & 0.420 & 0.508 \\
\rowcolor{gray!20} {+ARISE-RL}                & \textbf{0.479} & \textbf{0.781} & 0.523 & \textbf{0.643} & 0.373 & 0.127 & \textbf{0.587} & \textbf{0.812} & \textbf{0.797} & \textbf{0.342} & \textbf{0.456} & \textbf{0.569} \\
\bottomrule
\end{tabular}
}}
\end{table}

\section{Experiments}

\subsection{Experimental Settings}

\textbf{Datasets.}
We evaluate ARISE-RL on four open-ended agentic benchmarks. Two of them belong to our proposed ECR-Bench suite: {ECR-DeepResearch} for single-tool open-ended research and {ECR-Travel} for multi-tool travel planning, which is further decomposed into five subtasks---route planning (Direction), transportation comparison (Compare), nearby POI search (Search), one-day itinerary (1-Day), and multi-day itinerary (M-Day). The other two are existing rubric-based open-ended benchmarks: {ResearchRubrics}~\citep{sharma2025researchrubrics} for single-tool deep research and {VitaBench}~\citep{he2025vitabench} for multi-tool daily-life agents, covering three single-scenario domains, namely food delivery (Deliv.), in-store consumption (In-Store), and online travel services (OTA), as well as a cross-scenario setting (Cross) that combines tools from multiple domains within a single task. For the two single-tool research benchmarks, we report the rubric score rate; For the two multi-tool benchmarks, we report the task pass rate averaged across four independent runs.
\\
\textbf{Baselines.}
We compare ARISE-RL against three groups of competitive baselines, all evaluated under the same setting:
(i) \emph{closed-source LLMs}: Gemini3-Pro~\citep{google2025gemini3}, GPT-5~\citep{openai2025gpt5}, GPT-5.2~\citep{openai2025gpt52}, Claude-4.5-Sonnet~\citep{anthropic2025claude45} and Claude-4.6-Sonnet~\citep{anthropic2025claude46};
(ii) \emph{open-source LLMs}: Qwen3-32B, Qwen3-235B, and Qwen3.5-397B, which span two model generations and an order-of-magnitude range in parameter count;
(iii) \emph{self-evolution frameworks} built on the \emph{same} small backbones: Dr.~Zero~\citep{yue2026dr} and Absolute Zero~\citep{zhao2026absolute}, re-implemented under the identical rubric judge and tool stack.
\\
\textbf{Implementation Details.}
All experiments are implemented in PyTorch on 32 H20 GPUs. We sample $G{=}16$ rollouts per query at temperature $0.8$ and allow up to $40$ rounds of tool interaction. For the difficulty-shaped Generator reward, we set $K{=}8$ Solver rollouts per generated query and a success threshold $\gamma{=}0.9$. The Solver reward uses the partial-credit/full-completion weighting $(0.8, 0.2)$. For RG-SED we set the initial coefficient $\lambda_0^{a}{=}0.5$ for both roles, the activation threshold $\tau_a{=}0.05$. The UserSimulator that drives the interactive multi-turn benchmarks is instantiated with \texttt{Qwen3.5-397B}, and all judges, including rubric scoring, coach summarisation, and format checking, are uniformly served by \texttt{gpt-5.2}.

\begin{table}[!t]
\begin{minipage}[t]{0.48\textwidth}
\centering
\small
\caption{Ablation study on Qwen3.5-9B. “VitaBench” averages four sub-tasks (Deliv./In-Store/OTA/Cross), while “ECR-Travel” averages five sub-tasks.}
\label{tab:ablation}
\resizebox{\linewidth}{!}{%
\setlength{\tabcolsep}{4pt}
\renewcommand{\arraystretch}{1.05}
\begin{tabular}{l|cccc}
\toprule
\textbf{Variant} & \textbf{RR} & \textbf{ECR-DR} & \textbf{VitaBench} & \textbf{ECR-Travel} \\
\midrule
\textbf{ARISE-RL (full)}      & \textbf{0.479} & \textbf{0.781} & \textbf{0.417} & \textbf{0.599} \\
\midrule
w/o RG-SED                    & 0.422 & 0.716 & 0.350 & 0.518 \\
w/o reward gating             & 0.438 & 0.725 & 0.371 & 0.523 \\
w/o tool-grounded rubric      & 0.447 & 0.720 & 0.395 & 0.549 \\
w/o difficulty-shaped reward  & 0.456 & 0.743 & 0.402 & 0.537 \\
\bottomrule
\end{tabular}}
\end{minipage}\hfill
\begin{minipage}[t]{0.48\textwidth}
\centering
\small
\caption{Per-cycle Solver performance during the three co-evolution iterations of ARISE-RL on the Qwen3.5-9B backbone. Each cycle yields monotone improvement across all four benchmarks.}
\label{tab:iters}
\resizebox{\linewidth}{!}{%
\setlength{\tabcolsep}{4pt}
\renewcommand{\arraystretch}{1.05}
\begin{tabular}{l|cccc}
\toprule
\textbf{Cycle} & \textbf{RR} & \textbf{ECR-DR} & \textbf{VitaBench} & \textbf{ECR-Travel} \\
\midrule
Qwen3.5-9B  & 0.447 & 0.735 & 0.256 & 0.470 \\
\midrule
iter 1            & 0.458 & 0.752 & 0.310 & 0.512 \\
iter 2            & 0.470 & 0.768 & 0.367 & 0.561 \\
\textbf{iter 3 (final)} & \textbf{0.479} & \textbf{0.781} & \textbf{0.417} & \textbf{0.599} \\
\bottomrule
\end{tabular}}
\end{minipage}
\vspace{-3mm}
\end{table}

\subsection{Main Results}

\textbf{Comparison to Strong Baselines.}
Table~\ref{tab:main_results} reports the main results. ARISE-RL on the Qwen3.5-9B backbone consistently attains the best overall performance, surpassing all closed-source models and the strongest open-source non-self-evolving baselines, with the most pronounced gains on the two interactive multi-tool benchmarks while still reaching the column-wise best on the two rubric-based DeepResearch-style benchmarks. The advantage is preserved on the Qwen3-8B backbone. 
Under the same backbone, ARISE-RL consistently exhibits a clear performance advantage over two contemporary self-evolution baselines, Dr.~Zero and Absolute Zero. 
This result indicates that the synergy between RG-SED and the decoupled Generator--Solver design is the key factor behind the performance gains on long-horizon open-ended agentic tasks.
\\
\textbf{Comparison with On-Policy Distillation Methods.}
To isolate the contribution of RG-SED, we compare it against two representative on-policy distillation baselines, OPCD~\citep{ye2026policy} and GKD~\citep{agarwal2024policy}, re-implemented under the \emph{identical} Qwen3.5-9B backbone, training data, and rubric-judging pipeline.

\begin{figure}[!t]
  \centering
  \includegraphics[width=0.42\linewidth]{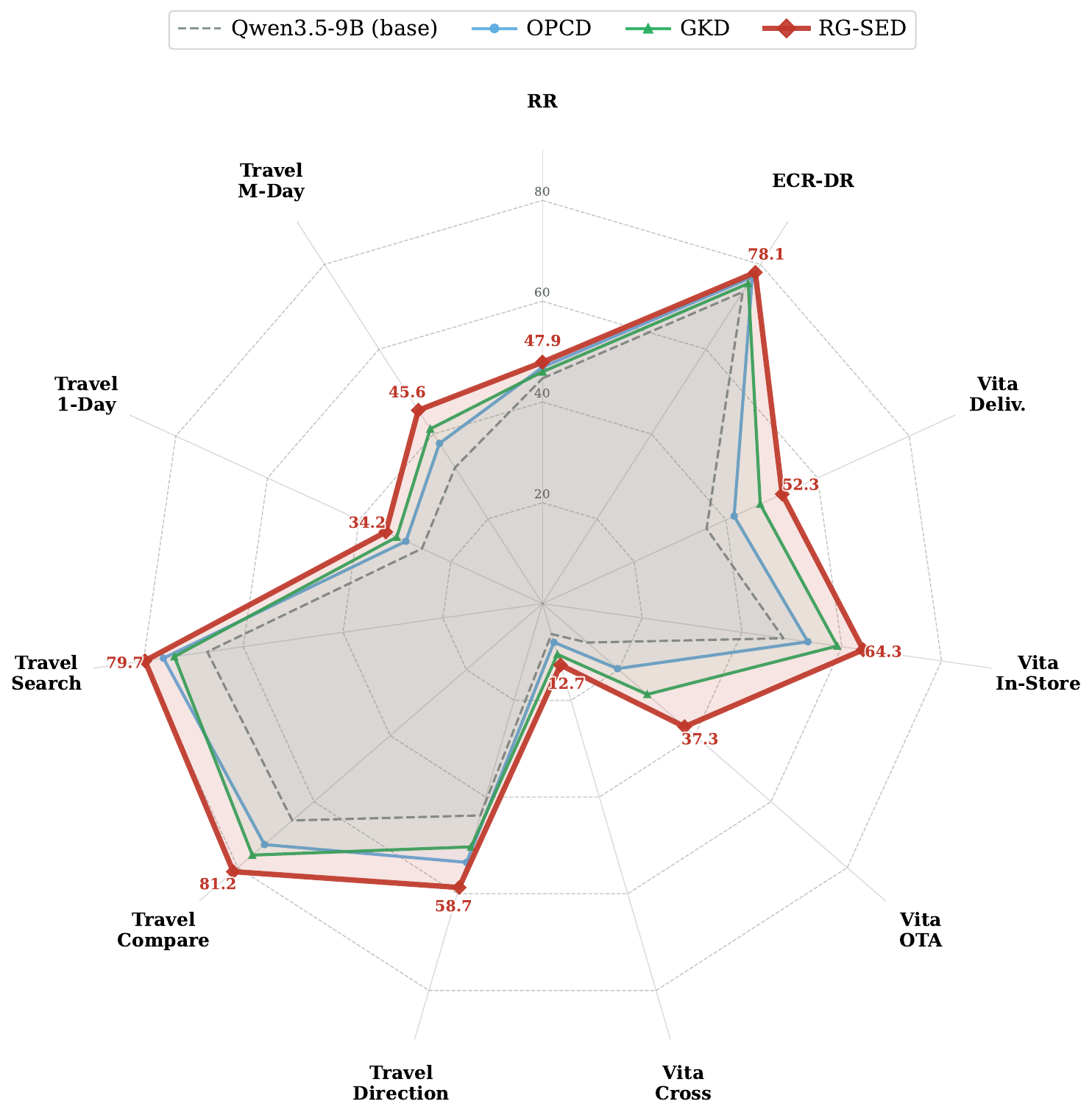}
  \caption{Detailed per-benchmark performance comparison of ARISE-RL, OPCD, and GKD.}
  \label{fig:radar}
  \vspace{-3mm}
\end{figure}
As shown in Figure~\ref{fig:radar}, RG-SED attains the best score on every one of the eleven benchmark axes.
Although OPCD and GKD exhibit complementary strengths on the radar, they remain consistently inferior to RG-SED overall. 
This confirms that our group-level memory-augmented self-distillation, in which the distillation signal is gated by empirical reward improvement rather than applied blindly, effectively enhances the learning capability of open-ended agents on complex tool-use tasks far beyond what a standard on-policy distillation loop can achieve independently.

\begin{figure}[!t]
\begin{minipage}[t]{0.48\textwidth}
  \centering
  \includegraphics[width=0.92\linewidth]{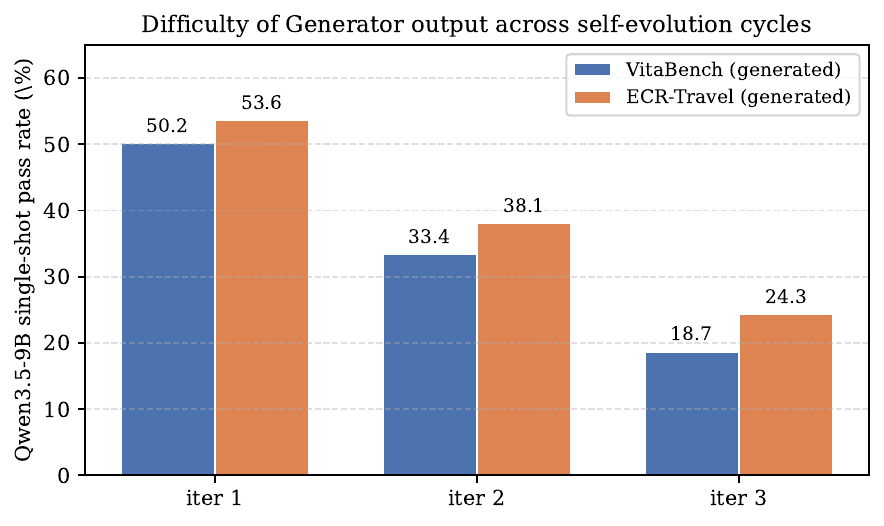}
  \caption{Single-shot pass rate of the \emph{frozen Qwen3.5-9B base policy} on generated questions sampled from the Generator at the end of each self-evolution cycle.}
  \label{fig:curriculum}
\end{minipage}\hfill
\begin{minipage}[t]{0.48\textwidth}
  \centering
  \includegraphics[width=0.92\linewidth]{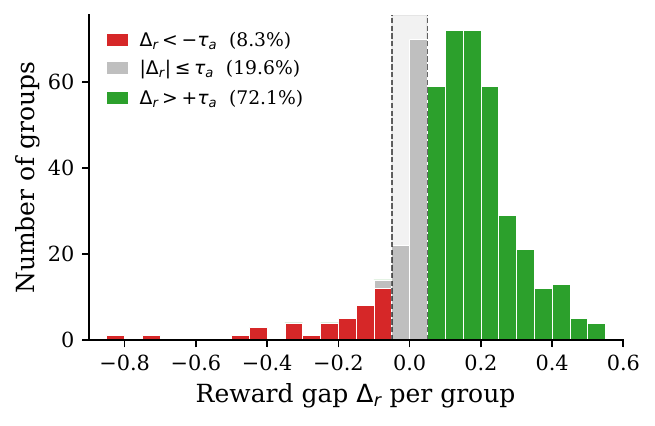}
  \caption{Reward-gap distribution on ECR-Travel. Per-group $\Delta_r$ over training groups, stacked by gate region.}
  \label{fig:gate-selectivity}
\end{minipage}
\vspace{-3mm}
\end{figure}

\subsection{Further Analysis}

\textbf{Ablation Study.}
Table~\ref{tab:ablation} reports the ablation results for four key design choices in ARISE-RL: \emph{w/o RG-SED}, \emph{w/o reward gating} (i.e., always applying the loss without the gate), \emph{w/o tool-grounded rubric}, and \emph{w/o difficulty-shaped reward}. RG-SED emerges as the most influential component, with the largest performance drop observed on the two multi-tool benchmarks, where reward-validated guidance is most critical. Within RG-SED, reward gating is itself essential: disabling it causes substantial degradation on multi-tool tasks, indicating that ungated distillation tends to amplify, rather than suppress, noisy supervision. The tool-grounded rubric and the difficulty-shaped reward each yield consistent gains, suggesting that the four design choices are complementary rather than redundant.
\\
\textbf{Self-Evolution Cycles.}
ARISE-RL is trained for three Generator--Solver co-evolution cycles. Table~\ref{tab:iters} shows that each cycle brings non-trivial performance gains, with the most pronounced improvements observed on the two multi-tool benchmarks. This trend is consistent with the design of ARISE-RL, where Solver improvement is accompanied by progressively harder yet still valid queries generated by the Generator.
Figure~\ref{fig:curriculum} further provides direct evidence that the generated queries become increasingly difficult over the course of self-evolution. Specifically, we evaluate the single-shot pass rate of the frozen Qwen3.5-9B base policy on questions freshly sampled from the Generator at the end of each self-evolution cycle. On both multi-tool benchmarks, the pass rate decreases monotonically and consistently as the Generator evolves across cycles. This result shows that the Generator is indeed producing increasingly challenging queries, thereby inducing an effective curriculum that matches the improving capability of the Solver.
\\
\textbf{Gate Selectivity on ECR-Travel.}
We evaluate the selectivity of the RG-SED reward gate on ECR-Travel. 
Figure~\ref{fig:gate-selectivity} shows the distribution of the reward gap $\Delta_r$ over all training groups. 
Among them, $8.3\%$ satisfy $\Delta_r<-\tau_a$, where coach memory reduces the Solver's empirical reward; $19.6\%$ fall into the ambiguous region $|\Delta_r|\le\tau_a$; and $72.1\%$ show a reliable positive gap. 
Thus, coach memory is not universally helpful: an ungated distillation objective would apply neutral or even harmful supervision to nearly one-third of the groups. 
The gate $\mathbb{I}[\Delta_r>\tau_a]$ prevents such negative transfer by activating distillation only when coach-augmented rollouts empirically improve reward.


\section{Conclusion}

We present ARISE-RL, a rubric-mediated co-evolution framework that couples a Generator and a Solver with fine-grained rewards, together with \emph{Reward-Gated Self-Evolution Distillation}, which distills memory-augmented behavior only when it empirically improves reward. We further introduce \emph{ECR-Bench}, an expert-calibrated rubric benchmark for multi-tool agents. Across four benchmarks, ARISE-RL achieves average state-of-the-art performance with a $9$B open-source backbone, demonstrating the effectiveness of rubric-mediated co-evolution and reward-gated self-distillation for open-ended agent training. 
This highlights the promise of unified closed-loop self-evolution for scalable open-ended agent training.

\section*{Limitations}

Due to computational resource constraints, all experiments are conducted on $8/9$B open-source backbone models. As a result, while ARISE-RL demonstrates consistent gains at this scale, the scalability of rubric-mediated co-evolution and RG-SED to substantially larger base models remains to be verified. We leave large-scale validation across stronger backbones to future work.

\section*{Ethical Considerations}
\label{app:ethics}

We will abide by the laws, rules, and regulations of our community, school, work, and country. We will conduct ourselves with integrity, fidelity, and honesty. We will openly take responsibility for our actions and only make agreements that we intend to keep. All data used in this study are intended for research purposes. No personally identifiable information (PII) was collected.
The dataset construction and data collection protocol have been reviewed and approved by our organization's internal Ethics Review Board, and all collected data are used solely for research purposes without collecting personally identifiable information.
For existing artifacts used in this work, including tools, benchmarks, APIs, and model resources, we follow their stated licenses, access conditions, and terms of use. Any artifacts created in this work, including benchmark data, rubrics, prompts, and evaluation scripts, are intended for research use only. Their use and distribution will be made consistent with the original access conditions of the underlying resources, and derivatives of research-only data will not be used outside research contexts.

\clearpage
{
\bibliography{custom}
\bibliographystyle{colm2024_conference}
}

\appendix
\newpage
\tcbset{
    promptbox/.style={
        colback=gray!5!white,
        colframe=gray!75!black,
        fonttitle=\bfseries,
        before upper=\ttfamily\scriptsize,
        listing options={
            basicstyle=\ttfamily\scriptsize,
            breaklines=true,
            columns=fullflexible,
            keepspaces=true,
        },
    }
}

\DefineVerbatimEnvironment{PromptVerb}{Verbatim}{%
    fontsize=\scriptsize,
    baselinestretch=0.95,
    breaklines=true,
    breakanywhere=true,
    breaksymbolleft={},
    breaksymbolright={},
}

\section{The Use of Large Language Models Statement}
\label{app:llm-use}

The authors use LLMs as an assistive tool in the preparation of this manuscript. We use LLMs to proofread, check grammar, and refine the language in the manuscript for improved clarity and readability.

\section{ECR-Bench Construction Details}
\label{app:ecr-bench}

This appendix complements Section~\ref{sec:ecr-bench} by spelling out (i) the tool inventory exposed to the agent on each domain, and (ii) descriptive statistics that characterise the resulting benchmark.

\subsection{Annotated Tool Inventory}
\label{app:ecr-bench:tools}

\paragraph{ECR-DeepResearch.} 
The agent is equipped with a single \texttt{web\_search} tool, implemented using the Google Custom Search API. 
To avoid context saturation from lengthy retrieved pages, we summarize any parsed page exceeding \(7{,}500\) characters with a dedicated Qwen3-Max summarizer before returning it to the agent. 






\paragraph{ECR-Travel.} ECR-Travel exposes a richer tool set of \textbf{seven} primitives:
\begin{itemize}
\setlength{\itemsep}{2pt}
\item \textbf{Poi\_search} --- a POI resolution tool built on Amap's place-search service. 
Given a structured address (e.g.\ ``\textit{No.~88 Century Avenue, Pudong New Area, Shanghai}'') or a named place query (e.g.\ ``\textit{West Lake Cultural Square}''), it returns ranked candidate POIs with complete addresses, geographic coordinates (\texttt{longitude, latitude}), and relevant business metadata.
\item \textbf{Around\_search} --- a radius-based nearby-place retrieval tool built on Amap. 
Given a center coordinate, a search radius, and optionally a POI type or keyword (e.g.\ ``\textit{coffee shop}''), it returns nearby POI candidates using the same output schema as \texttt{poi\_search}.

\item \textbf{Direction} --- a multi-modal routing tool built on Amap navigation services. 
Given origin and destination coordinates, with optional waypoints and a travel-mode flag, it returns a step-by-step route plan for walking, driving, or public transit, including distance and estimated duration for each leg.

\item \textbf{Web\_search} --- a general-purpose textual search tool implemented with the Google Custom Search API. 
For open-ended travel questions that cannot be fully answered by structured map APIs alone, such as ``\textit{what are some quiet places to visit in Hangzhou?}'' or ``\textit{which evening activities are suitable for families in Chengdu?}'', the agent uses this tool to obtain prose-style background information and recommendations. 
As in ECR-DeepResearch, retrieved pages longer than \(7{,}500\) characters are summarized in-line before being returned to the agent.

\item \textbf{Search\_flights} --- a date-specific intercity flight search tool. 
Given a departure city, an arrival city, and a travel date, it returns ranked flight options with flight numbers, fares, departure and arrival airports, and scheduled times.

\item \textbf{Search\_train\_tickets} --- a date-specific intercity train-ticket search tool. 
Given a departure city, an arrival city, and a travel date, it returns ranked train options with train IDs, fares, departure and arrival stations, scheduled times, and whether the itinerary is direct or requires transfer.

\item \textbf{Weather} --- a city-level weather lookup tool for historical and forecast conditions. 
Given a city and a date range, it returns daily weather records, including weather conditions, daytime and nighttime temperatures, wind, and humidity. 
Unlike prior Amap-only travel benchmarks, this tool enables the agent to ground schedule planning in real weather conditions.
\end{itemize}
\texttt{poi\_search}, \texttt{around\_search}, \texttt{direction}, and \texttt{weather} are wrappers over Amap's Web Service API. \texttt{web\_search} is implemented with Google Custom Search, while \texttt{search\_flights} and \texttt{search\_train\_tickets} are simulated by GPT-5-mini under a deterministic prompt that encodes a closed database of flights and trains for a fixed planning horizon.

\subsection{Benchmark Statistics}
\label{app:ecr-bench:stats}

\begin{figure}[!t]
  \centering
  \includegraphics[width=0.9\linewidth]{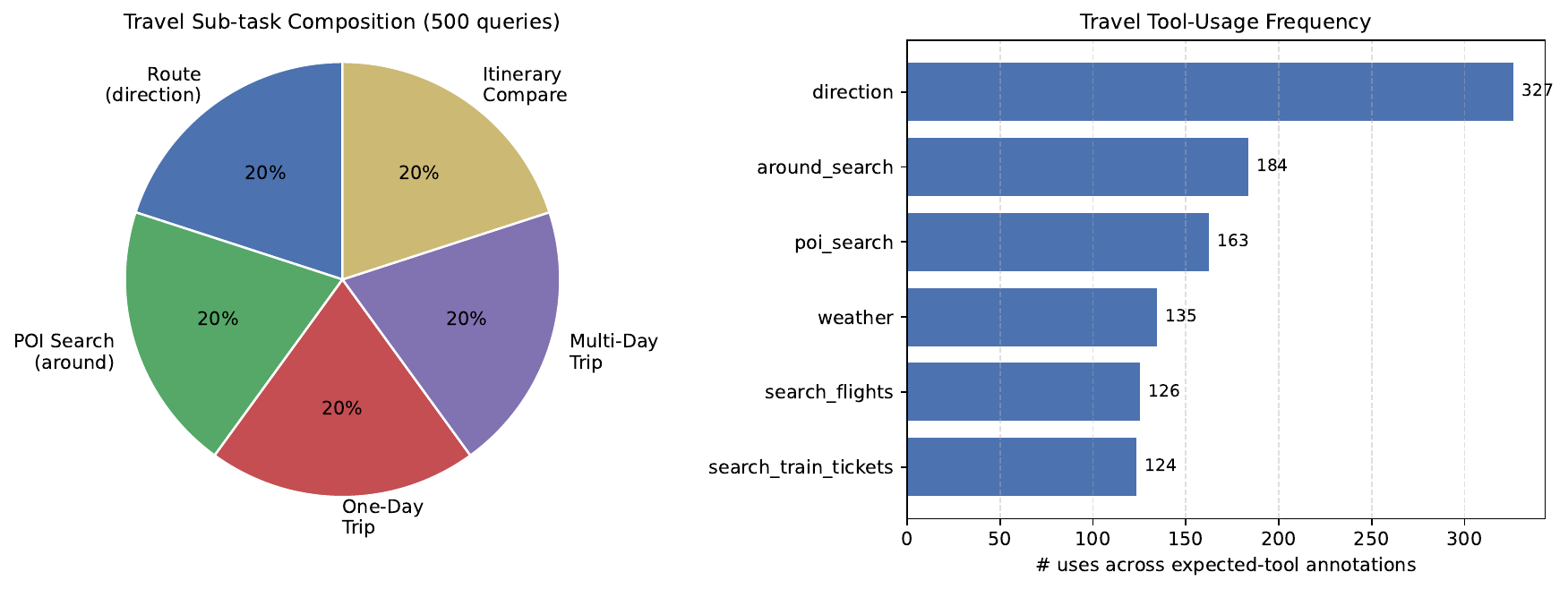}
  \caption{\textbf{Left:} the five ECR-Travel sub-tasks are perfectly balanced (100 queries each). \textbf{Right:} aggregated tool-usage frequency across all annotated \texttt{expected\_tools}; \texttt{direction} dominates, while \texttt{weather} and the transport-search tools jointly account for nearly a third of all expected calls.}
  \label{fig:travel-task-tools}
\end{figure}

\paragraph{Sub-task and tool coverage (ECR-Travel).} Figure~\ref{fig:travel-task-tools} summarises the sub-task composition and the underlying tool-usage frequency across all 500 ECR-Travel queries' annotated \texttt{expected\_tools}, i.e., rubric items specifying the required tool-use behavior. The five sub-tasks are perfectly balanced (100 queries each), and \texttt{direction} is the most frequently invoked primitive ($327$ uses), followed by \texttt{around\_search} ($184$), \texttt{poi\_search} ($163$), \texttt{weather} ($135$), and the two transport-search tools at $\approx 125$ uses each. 

\begin{figure}[!t]
\begin{minipage}[t]{0.48\textwidth}
  \centering
  \includegraphics[width=\linewidth]{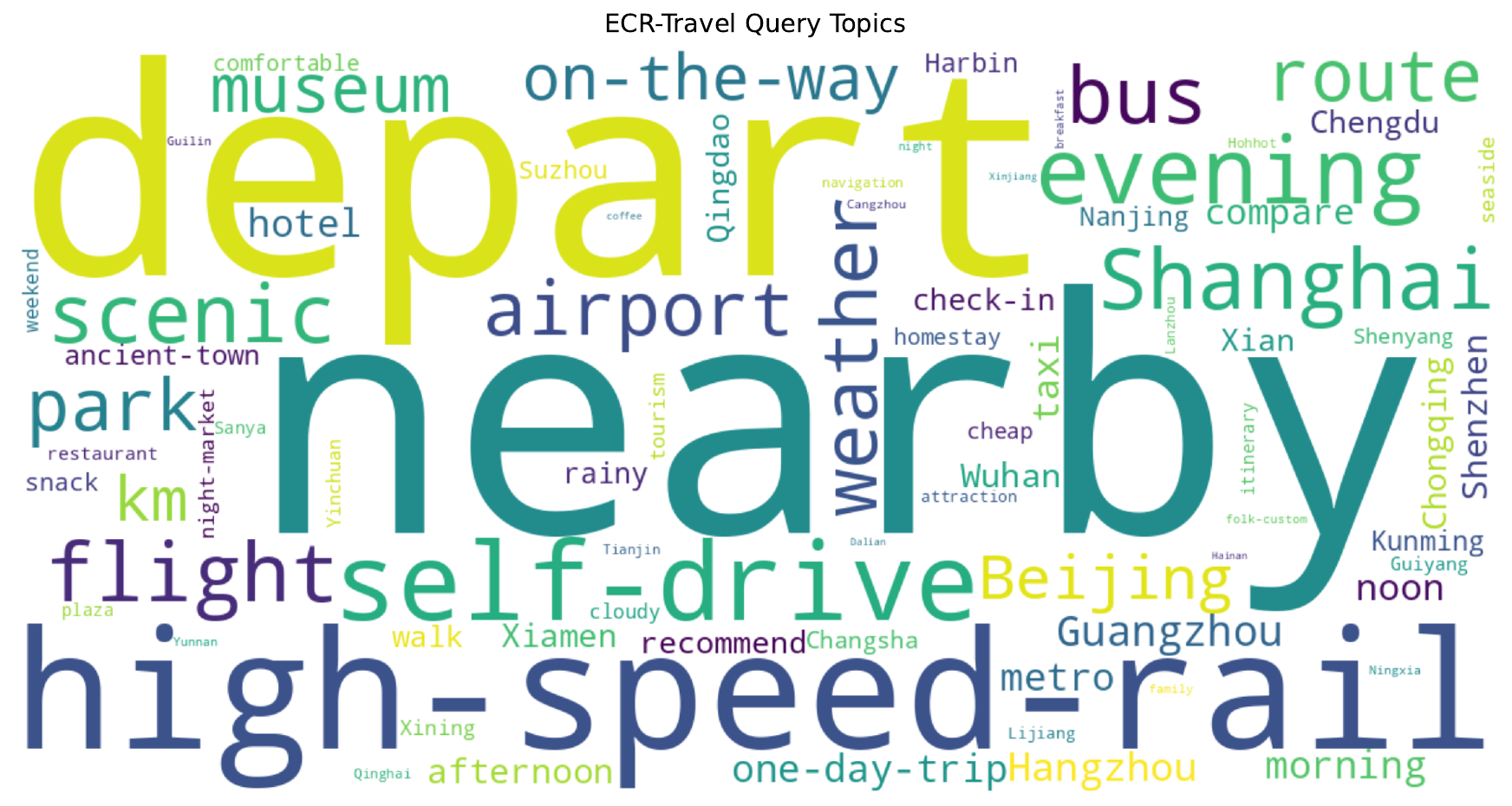}
  \caption{Word cloud of ECR-Travel query topics. Frequencies are computed on jieba tokens whose Chinese form is mapped to English via a curated travel lexicon (cities, activities, transport, time-of-day, weather).}
  \label{fig:wc-travel}
\end{minipage}\hfill
\begin{minipage}[t]{0.48\textwidth}
  \centering
  \includegraphics[width=\linewidth]{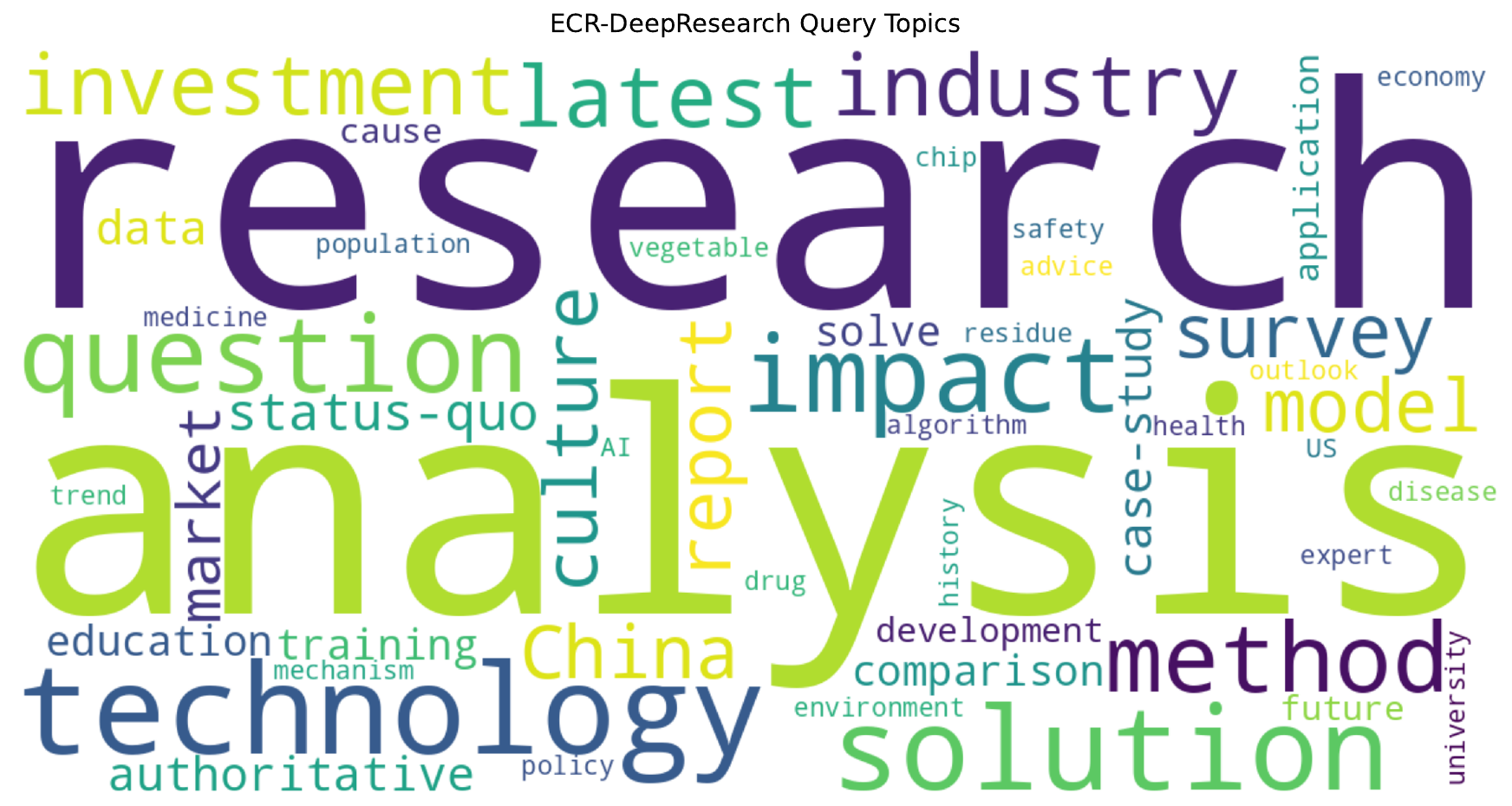}
  \caption{Word cloud of ECR-DeepResearch query topics. The flat topic distribution illustrates the open-domain nature of the benchmark: no single research theme dominates.}
  \label{fig:wc-dr}
\end{minipage}
\end{figure}

\paragraph{Topic diversity.} Figure~\ref{fig:wc-travel} and Figure~\ref{fig:wc-dr} show word clouds built from jieba-segmented query tokens, after mapping the top frequent Chinese terms to English; research/health/technology/finance domains on the DR side. ECR-Travel concentrates around concrete travel verbs and modes (\textit{depart}, \textit{nearby}, \textit{self-drive}, \textit{high-speed-rail}, \textit{flight}, \textit{evening}) and a wide set of Chinese cities. ECR-DeepResearch instead spreads across a much flatter topic distribution that includes \textit{technology}, \textit{health}, \textit{economy}, \textit{policy}, and \textit{environment}, confirming the open-domain nature of the benchmark.

\begin{figure}[!t]
  \centering
  \includegraphics[width=0.72\linewidth]{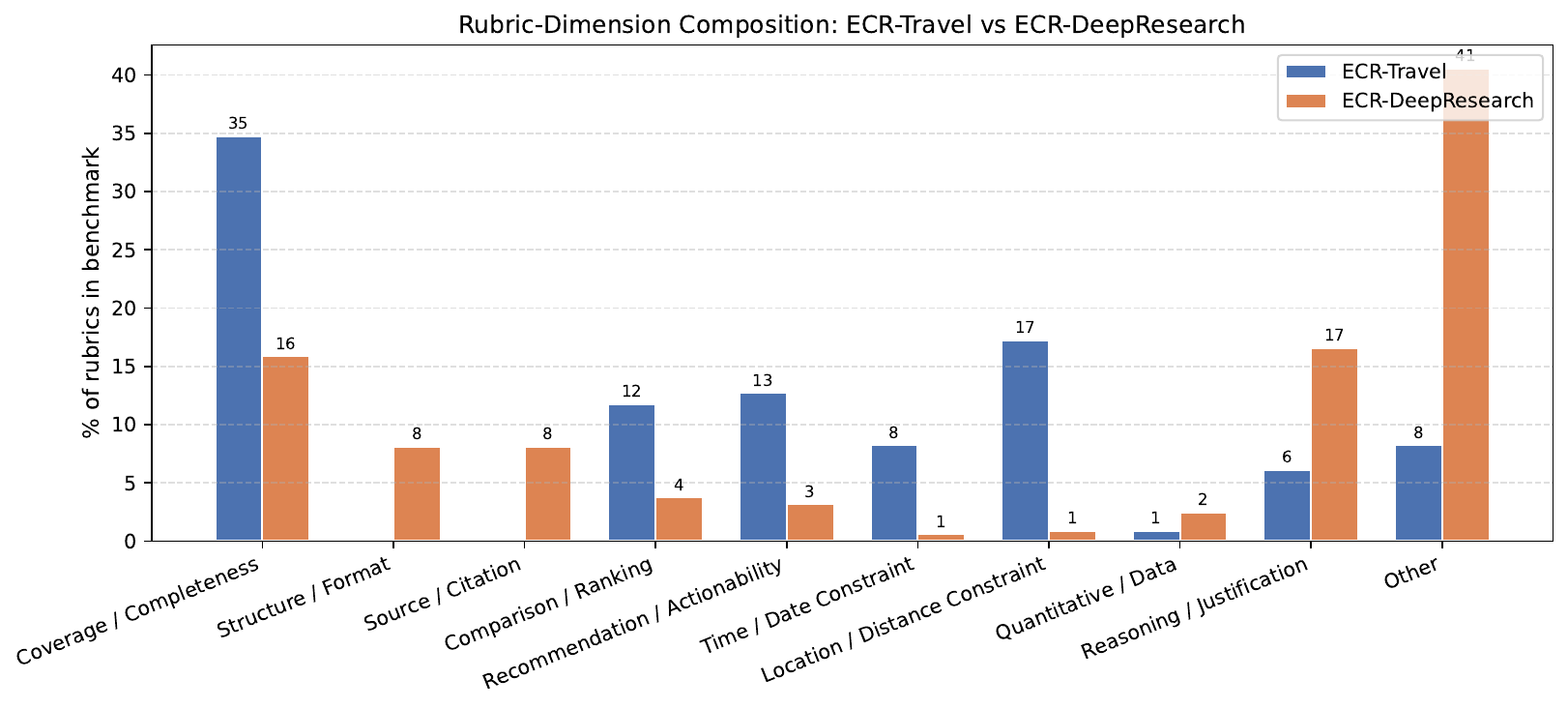}
  \caption{Composition of rubrics by content dimension. ECR-Travel is dominated by coverage + space/time constraints; ECR-DeepResearch is denser on reasoning, sourcing, and structural formatting. The \emph{Other} bucket collects rubrics whose surface form did not match any of the nine keyword patterns, and that mostly correspond to free-form ``answer is logically consistent'' style criteria.}
  \label{fig:rubric-dims}
\end{figure}

\paragraph{Rubric-dimension composition.} Each rubric was assigned to one of ten dimensions by keyword pattern matching (\emph{coverage}, \emph{structure}, \emph{source}, \emph{comparison}, \emph{recommendation}, \emph{time}, \emph{location}, \emph{quantitative}, \emph{reasoning}, \emph{other}). Figure~\ref{fig:rubric-dims} compares the per-bench distribution. ECR-Travel rubrics are heavily skewed toward \emph{coverage} (whether the agent's reply mentions all expected POIs / sub-tasks) and \emph{location-} / \emph{time-} constraints (whether the answer respects the geographical and temporal envelope of the trip), reflecting the agentic nature of the task. ECR-DeepResearch rubrics are more evenly spread across \emph{coverage}, \emph{reasoning}, and a long tail of presentation-style rubrics (\emph{source}, \emph{structure}, \emph{quantitative}), reflecting that a deep-research answer is scored not only on what it says but on how the conclusion is supported and laid out.

\subsection{Instructions Given to Rubric Annotators}
\label{app:annotator-instructions}

Rubrics were authored by $6$ domain experts ($3$ in travel planning and $3$ in deep-research / academic writing). All annotators were informed of the purpose and scope of the study and agreed to participate. Each annotator received a standardized bilingual (Chinese--English) annotation handbook covering: (i) task background and the intended agent behaviour for each query; (ii) rubric-writing guidelines (each rubric must be an independently verifiable single assertion with an explicit pass criterion); (iii) the semantics of the scoring scale with worked examples; (iv) the format of tool-call rubrics, including \texttt{expected\_tools} examples; and (v) a risk disclaimer stating that annotations are used solely for academic research, that no personally identifiable information about the annotators is disclosed, and that participation can be withdrawn at any time. Annotation was conducted by in-house domain colleagues as part of their research collaboration, with no external compensation involved.

\section{Baseline Details}
\label{app:baseline-details}

To complement the results in Table~\ref{tab:main_results} and Figure~\ref{fig:radar}, we organize the evaluated baselines into four groups: closed-source LLMs, open-source backbone models, self-evolution frameworks built on the same backbones, and on-policy distillation methods.
All baselines are evaluated through the same benchmark interface, tool environment, and scoring pipeline; the on-policy distillation methods are used only for the RG-SED mechanism comparison in Figure~\ref{fig:radar}.

\paragraph{Closed-source LLMs.}
\begin{itemize}
\setlength{\itemsep}{2pt}
  \item \textbf{Gemini3-Pro}~\citep{google2025gemini3}: a proprietary general-purpose model from Google DeepMind, included as a strong commercial agent baseline across all evaluated benchmarks.
  \item \textbf{GPT-5}~\citep{openai2025gpt5}: a proprietary OpenAI model, used to measure how ARISE-RL compares with a commercial general-purpose LLM under the same agent interface.
  \item \textbf{GPT-5.2}~\citep{openai2025gpt52}: a later proprietary OpenAI model in the GPT-5 family, included to test whether a stronger commercial model directly solves the open-ended agentic benchmarks.
  \item \textbf{Claude-4.5-Sonnet}~\citep{anthropic2025claude45}: a proprietary Anthropic Sonnet model, providing a commercial model family for comparison.
  \item \textbf{Claude-4.6-Sonnet}~\citep{anthropic2025claude46}: a later proprietary Sonnet model, used to evaluate a stronger Claude-family baseline on the tool-use benchmarks.
\end{itemize}

\begin{figure}[!t]
  \centering
  \includegraphics[width=1\linewidth]{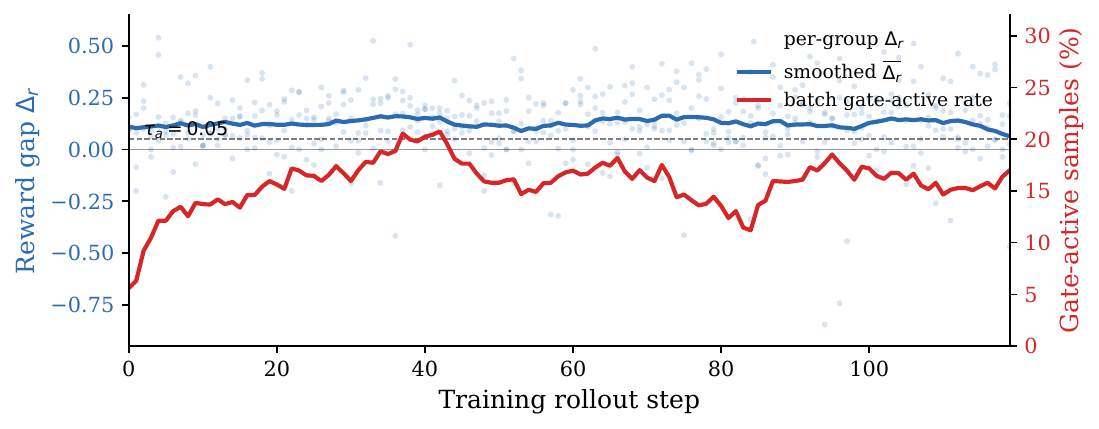}
  \caption{\textbf{RG-SED gate dynamics on ECR-Travel.} Per-group $\Delta_r$ (light blue scatter) and its smoothed mean $\overline{\Delta_r}$ (blue) stay above $\tau_a$ throughout training, while the gate-active sample rate per batch (red, right axis) ramps to $\approx 22\%$ during warm-up and is then modulated in real time by the running $\Delta_r$ distribution---most visibly retracting around rollouts $80$--$90$ where coach memory degrades. Selectivity is therefore adaptive and emergent, not scheduled.}
  \label{fig:gate-curriculum}
\end{figure}

\paragraph{Open-source LLMs.}
\begin{itemize}
\setlength{\itemsep}{2pt}
  \item \textbf{Qwen3-32B}~\citep{yang2025qwen3}: a mid-scale open-source Qwen backbone, used as a non-self-evolving open-source baseline.
    \item \textbf{Qwen3-235B}~\citep{yang2025qwen3}: a large Mixture-of-Experts Qwen3 backbone with 235B total parameters and 22B activated parameters, included to measure the gain from parameter scaling without self-evolution.
    \item \textbf{Qwen3.5-397B}~\citep{qwen2026qwen35}: a large-scale Mixture-of-Experts model from the Qwen3.5 series, with 397B total parameters and 17B activated parameters.
  \item \textbf{Qwen3-8B}~\citep{yang2025qwen3}: a smaller open-source base policy used for same-backbone comparisons among Dr.~Zero, Absolute Zero, and ARISE-RL.
  \item \textbf{Qwen3.5-9B}~\citep{qwen2026qwen35}: the main 9B-scale open-source base policy used for the principal ARISE-RL run, same-backbone self-evolution baselines, and RG-SED comparisons.
\end{itemize}

\paragraph{Self-Evolving Frameworks.}
\begin{itemize}
\setlength{\itemsep}{2pt}
  \item \textbf{Dr.~Zero}~\citep{yue2026dr}: a self-evolving search-agent framework without human-authored training data, re-implemented on the same small backbones, tool environment, and rubric judge.
  \item \textbf{Absolute Zero}~\citep{zhao2026absolute}: a zero-external-data reinforced self-play reasoning framework, adapted as a same-backbone self-evolution baseline. Following its self-play design, we instantiate the task Generator and Solver with the same backbone model.
\end{itemize}

\paragraph{On-policy distillation methods.}
\begin{itemize}
\setlength{\itemsep}{2pt}
  \item \textbf{OPCD}~\citep{ye2026policy}: On-Policy Context Distillation, used to compare RG-SED with context-distillation-based on-policy learning under the same Qwen3.5-9B setup. The external teacher is set to Qwen3.5-397B.
  \item \textbf{GKD}~\citep{agarwal2024policy}: an on-policy distillation baseline. The external teacher is set to Qwen3.5-397B.
\end{itemize}

\section{More Experiments}
\subsection{Gate Dynamics over Training.}
We further trace how $\Delta_r$ and the gate-active rate co-evolve along the rollout axis (Figure~\ref{fig:gate-curriculum}). The smoothed mean $\overline{\Delta_r}$ stays above the activation threshold $\tau_a$ throughout training, while per-group $\Delta_r$ retains substantial dispersion (light blue scatter), indicating that coach memory continues to deliver high-variance but on-average positive guidance even as the Solver strengthens. The fraction of batch samples in which the gate fires (red, right axis) climbs from $\approx 5\%$ to $\approx 22\%$ over the first $\sim 40$ rollouts under the cosine warm-up schedule, and \emph{does not} subsequently flatten: it is modulated in real time by the $\Delta_r$ distribution. The visible dip around rollouts $80$--$90$ coincides with a cluster of queries on which the coach yields a degraded $\Delta_r$, and the gate quietly retracts the distillation pressure rather than amplifying the noise. This adaptive, self-paced selectivity is not the product of any hand-tuned schedule but an emergent effect of pairing the gate with empirical reward feedback, mirroring the \emph{w/o reward gating} degradation in Table~\ref{tab:ablation}: distillation stays well-conditioned only when the gate retracts in real time as $\Delta_r$ dictates.

\begin{figure}[!t]
  \centering
  \includegraphics[width=0.5\linewidth]{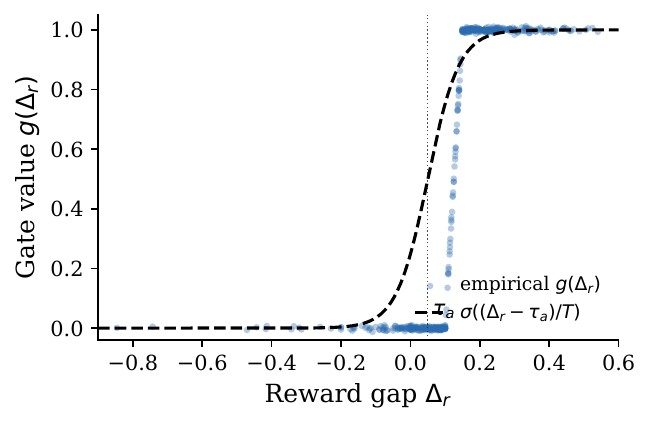}
  \caption{\textbf{Empirical RG-SED gate response on ECR-Travel.} For every training group, the empirical gate value $g(\Delta_r)$ tightly tracks the theoretical sigmoid $\sigma((\Delta_r-\tau_a)/T)$, confirming that the gate behaviour does not drift over training.}
  \label{fig:gate-response}
\end{figure}

\subsection{Empirical Gate Response.}
To further verify the stability of the reward gate, we overlay the empirical gate value $g(\Delta_r)$ produced by each training group against the theoretical sigmoid $\sigma\!\big((\Delta_r-\tau_a)/T\big)$ (Figure~\ref{fig:gate-response}). 
Across the entire ECR-Travel run, all empirical points fall tightly on the closed-form curve with no observable drift in later training. 
This shows that the gate reliably performs its intended role by selectively amplifying the distillation signal according to $\Delta_r$ throughout training, without requiring additional calibration.

\subsection{Distribution Divergence.}
\begin{figure}[!t]
  \centering
  \includegraphics[width=0.55\linewidth]{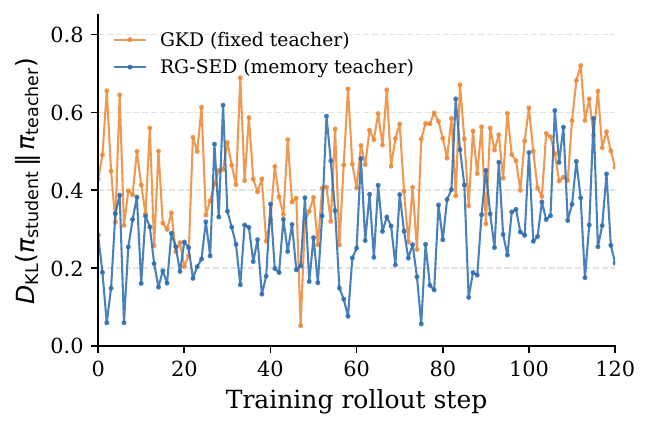}
  \caption{\textbf{Student--teacher distribution divergence on ECR-Travel.} Token-level $D_{\mathrm{KL}}(\pi_{\mathrm{student}}\Vert\pi_{\mathrm{teacher}})$ per training rollout step (mean and $\pm 1\sigma$ band across three seeds). Both runs fluctuate and cross at several rollout segments. Still, the running level under RG-SED's memory teacher is generally lower than under GKD's fixed teacher, consistent with the design intuition that a same-policy teacher upper-bounds the action-distribution gap by the prompt-induced shift.}
  \label{fig:gate-vs-gkd-kl}
\end{figure}

We empirically check the design assumption that RG-SED naturally bounds the student--teacher distribution gap by deriving the teacher from the \emph{same} policy under a different prompt. 
We track per-step token-level $D_{\mathrm{KL}}(\pi_{\mathrm{student}}\Vert\pi_{\mathrm{teacher}})$ during ECR-Travel training and compare RG-SED with GKD~\citep{agarwal2024policy}, a representative on-policy distillation baseline whose teacher is fixed and external. 
The external teacher is set to Qwen3.5-397B. 
As shown in Figure~\ref{fig:gate-vs-gkd-kl}, both curves exhibit the usual training-time fluctuations, and the two runs cross at several rollout segments. 
Across the run as a whole, however, RG-SED tends to operate at a lower divergence level, typically in the $\approx 0.2$ to $0.35$ range with only occasional brief excursions, while GKD remains in a higher band of $\approx 0.4$ to $0.55$ as the student drifts further from the fixed teacher under RL updates. 
This pattern is qualitatively consistent with the same-policy memory-teacher construction. 
Because student and teacher share the underlying parameters and differ only through prompt-induced shifts, the action-distribution gap is intrinsically constrained, without requiring additional explicit regularization to prevent it from growing unbounded.

\subsection{LLM--Human Judge Consistency.}
\begin{figure}[!t]
  \centering
  \includegraphics[width=0.45\linewidth]{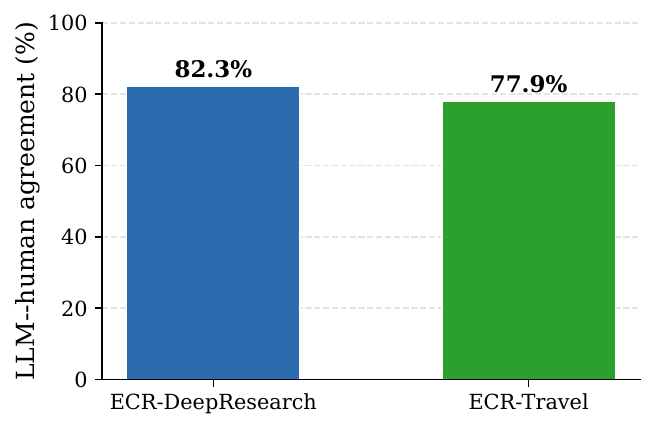}
  \caption{LLM judge vs.\ human agreement on ECR-DeepResearch and ECR-Travel.}
  \label{fig:judge-consistency}
\end{figure}

To assess the reliability of the LLM-based evaluation mechanism, we conduct a human-validation study on both ECR-DeepResearch and ECR-Travel. For each benchmark, we randomly sample $50$ final-model responses, score them with gpt-5.2 as the LLM judge under the rubric, and have three domain-familiar annotators independently score the same responses using the \emph{identical} rubric, taking the majority vote as the human label. Figure~\ref{fig:judge-consistency} reports the overall agreement between the gpt-5.2 judge and the human label per benchmark: $82.3\%$ on ECR-DeepResearch and $77.9\%$ on ECR-Travel. This relatively high level of consistency reflects improvements that are broadly aligned with human assessment.

\subsection{Generator Query Generation Pipeline}
\label{app:gen-pipeline}

We conduct three rounds of Generator--Solver evolution, refreshing the training pool every $120$ training steps. 
At each refresh, we pause RL updates and prompt the Generator to produce a new set of queries for the next training phase.
For each prompt, we perform a single rollout, extract the contents of the \texttt{<question>} and \texttt{<rubric>} tags, and retain $500$ non-duplicate queries after strict deduplication. 
To further reduce query-mode homogenization, we include the \emph{ten most recently generated queries} as in-context negatives in the Generator's system prompt and explicitly instruct the Generator to avoid duplicating or generating topically near-duplicate queries. 
This lightweight pipeline, which combines history-aware prompting with post-hoc deduplication, preserves the coverage and diversity of the training distribution without introducing additional training overhead.

\section{Prompt Templates}
\label{app:prompt}


This appendix lists the prompts used in ARISE-RL. We group them by role: \textbf{Generator} (task/rubric author), \textbf{Solver} (agent that completes the task), and \textbf{Judge} (rubric-based scoring LLM). For each role, we additionally include the \textit{coach memory} prompt that summarizes group-level feedback into a role-specific memory used by RG-SED. All prompts are translated into English here; the production system runs them in the language native to each task (Chinese for VitaBench / ECR-Travel; bilingual for ECR-DeepResearch / ResearchRubrics). Specifically, \texttt{expected\_tools} denotes the rubric item that specifies the required tool-call behavior. 
These prompts define the interaction protocol among task generation, tool-augmented solving, rubric-based judging, and memory-guided self-distillation.

\subsection{Generator Prompts}
\label{app:prompt:generator}

The Generator authors a task (query + rubrics) for the Solver. It must first call on tools to ground every rubric in real observations. The three task-specific system prompts are shown in Figures~\ref{fig:gen-dr}, \ref{fig:gen-travel}, and \ref{fig:gen-vita}; the shared coach-memory prompt that distills group-level feedback into a Generator-side memory is shown in Figure~\ref{fig:gen-coach}.

\begin{figure}[!ht]
\begin{tcolorbox}[promptbox, title=Generator System Prompt --- ECR-DeepResearch / ResearchRubrics]
\begin{PromptVerb}
You are a deep-research question generator. Your job is to author high-quality, diverse research-style questions for the solver.

## Core Principles
1. You MUST first call `web_search` to gather real data; never fabricate a question out of thin air. Every rubric must be supported by evidence found in the search results.
2. Diversify your search strategy: probe the topic from different angles, starting from overview queries and drilling down to specific facts and the latest developments.
3. Each rubric is a true/false assertion that the solver's final report can be checked against. Pay attention to data points, names, organizations and dates in the search results.
4. At most one `web_search` call per turn; plan ahead and explore over multiple turns.

## Workflow
Step 1 -- Explore the topic via `web_search` (2-5 queries with different keywords).
Step 2 -- Pick a research angle that genuinely requires multi-hop searching to answer and design one natural-sounding question.
Step 3 -- Design rubrics covering information coverage, structured presentation, source traceability and term accuracy. Rubrics must be tied to evidence in the search results and must not be overly strict.

## Output Format (strict JSON)
{
  "query": "<the generated research question>",
  "rubrics": [
    "rubric 1: a concrete verifiable assertion",
    "rubric 2: another verifiable assertion"
  ]
}
\end{PromptVerb}
\end{tcolorbox}
\caption{Generator system prompt for ECR-DeepResearch / ResearchRubrics. The Generator must call \texttt{web\_search} before composing each question, and every rubric must be tied to evidence in the search results.}
\label{fig:gen-dr}
\end{figure}

\begin{figure}[!ht]
\begin{tcolorbox}[promptbox, title=Generator System Prompt --- ECR-Travel]
\begin{PromptVerb}
You are a travel-planning question generator. Your job is to author high-quality, diverse travel-planning questions for the solver.

## Core Principles
1. Always call tools first to obtain real data; never fabricate a question.
2. Every tool listed in `expected_tools` must have been called by you and returned a valid result.
3. Broaden the question pool with `web_search` and `poi_search`: - explore arbitrary themes (events, festivals, foods, etc.); - blend the real information you find (place names, local specialties) into the question to make it grounded; - use map / transport tools to obtain concrete data (routes, weather, prices) for multi-tool tasks.
4. At most 3 tool calls per turn; split into multiple turns if needed.
5. Cover diverse cities and scenarios; differ noticeably from the seed examples.

## Workflow
Step 1 -- Call tools to gather data appropriate to the task type. Start with `web_search` for inspiration, then map / transport tools for the concrete numbers.
Step 2 -- Generate one natural, fluent question grounded in the tool results.
Step 3 -- Design rubrics, each a true/false assertion focused ONLY on what the user explicitly asks for. Do NOT add implicit requirements.

## Question Requirements
- Read like something a real user would ask.
- Do not stack many constraints in a single question.
- Use natural language; avoid symbols such as `->`, `/`, `vs`, `|` for routes or comparisons -- spell them out as "from A to B",
  "compared with", "or", etc.

## expected_tools Requirements
- Only include tools you successfully called.
- Arguments must use real values returned by tools (real place names, real coordinates). - `direction` tools: origin must differ from destination. - `weather`: include only if the question mentions weather.

## Output Format (strict JSON)
{
  "query": "<the generated question>",
  "expected_tools": [
    {"name": "<tool>", "arguments": {"<arg>": "<real value>"}}
  ],
  "rubrics": [
    "rubric 1: a concrete verifiable assertion",
    "rubric 2: another verifiable assertion"
  ]
}
\end{PromptVerb}
\end{tcolorbox}
\caption{Generator system prompt for ECR-Travel. The Generator must verify every entry in \texttt{expected\_tools} by calling it first, and rubrics are grounded in real POI / route / weather data returned by the tools.}
\label{fig:gen-travel}
\end{figure}

\begin{figure}[!ht]
\begin{tcolorbox}[promptbox, title=Generator System Prompt --- VitaBench]
\begin{PromptVerb}
You are an expert task author for VitaBench (OTA / daily-life services in an interactive setting).

Interactive setting characteristics:
- A UserSimulator gradually reveals the needs in your `instructions` to the agent across multiple turns.
- The agent may ask clarifying questions before calling tools.

## Workflow
Phase 1 -- Explore the environment by calling real OTA tools: * search hotels, flights, trains and POIs in the target city; * query weather; fetch detailed product info (room types, seats, ...). * The environment is a closed simulation, NOT the real internet: stick to the cities/routes available; if a search returns empty, change direction instead of retrying.

Phase 2 -- Design an interactive task based on the real data:
1. Write `instructions` in the second person ("you"), describing the user's situation and needs with enough context so that the UserSimulator can disclose information across multiple turns.
2. Design rubrics: each is a true/false assertion covering the key decision points (which hotel, which room, dates, paid or not). Use rubrics, focused on explicit needs only.

Phase 3 -- Emit JSON.

## Core Design Principles - `instructions` must be grounded in tool results.
- IDs, prices and room types in rubrics must come from tool returns.
- Exploit distractor entities: design conditions that pick out one specific option from several.
- Include date / time reasoning where appropriate.

## Output Format (strict JSON)
{
  "instructions": "<task instruction in second person>",
  "rubrics": ["rubric 1", "rubric 2", ...]
}

Note: do NOT output `environment` or `user_profile`; the system handles them automatically.
\end{PromptVerb}
\end{tcolorbox}
\caption{Generator system prompt for VitaBench. The Generator explores the closed simulation through real OTA tools, then composes interactive-mode instructions whose IDs / prices / dates are pulled from the actual tool returns.}
\label{fig:gen-vita}
\end{figure}

\begin{figure}[!ht]
\begin{tcolorbox}[promptbox, title=Generator Coach Memory Prompt (representative; one variant per task)]
\begin{PromptVerb}
You are a coach for the question-generation training loop. Below are {N} generated questions in the current group.

## Reward Semantics - reward = 0: the generator did not call tools, or the output was malformed. - reward in (0, peak): the difficulty drifts away from K/2 but still produces useful contrast. - reward = peak: the question lands on the solver's capability boundary (~K/2 of K trials succeed). Reward decays linearly to 0 at both ends ("triangle" reward).

## Current Group
- Average reward: {avg_reward}
- Reward distribution: {reward_dist}

{per_sample_summaries: index, reward, brief query, #rubrics, ...}

## Produce Coaching Advice for the Next Round (<= 300 words)
1. Diagnosis: What is the dominant failure mode in this group?
2. Diversity: Which domains/scenarios/angles should be explored next?
3. Difficulty: how to push the question closer to the K/2 peak -- harder rubrics, tighter constraints, or vice versa?
4. Concrete suggestion: give 1-2 specific question directions (topic + angle + suggested tool combination).

Output the advice directly; do not echo the inputs.
\end{PromptVerb}
\end{tcolorbox}
\caption{Coach-memory prompt for the Generator. The coach LLM summarizes a group of $N$ recent questions and their downstream reward signals into a short, role-conditioned memory that is injected at the tail of the Generator system prompt on the next rollout group.}
\label{fig:gen-coach}
\end{figure}

\subsection{Solver Prompts}
\label{app:prompt:solver}


The Solver answers the task generated upstream. Coach feedback (when active) is appended to the system prompt at training time. The three task-specific solver prompts are shown in Figures~\ref{fig:sol-dr}, \ref{fig:sol-travel}, and \ref{fig:sol-vita}; the corresponding Solver-side coach-memory prompt is shown in Figure~\ref{fig:sol-coach}.

\begin{figure}[!ht]
\begin{tcolorbox}[promptbox, title=Solver System Prompt --- ECR-DeepResearch / ResearchRubrics]
\begin{PromptVerb}
# Deep Research Specification You are a professional deep-research assistant. For the user's research question you must collect comprehensive, accurate information through multiple rounds of `web_search` and produce a high-quality research report.

## Research Workflow
1. Decompose: split the question into searchable sub-questions.
2. Multi-round search: each round focuses on one sub-question with precise keywords.
3. Verify: cross-check key facts using different keywords.
4. Synthesize: produce a structured report citing sources.

## Tool Use - {step_idx} of {max_steps} tool-call rounds used so far.
- You MUST call tools to query real data before answering.
- Each search should have a clear purpose; avoid duplicate searches.
- Keep keywords specific; avoid being too broad.

## Answer Quality
- Comprehensive: cover every aspect of the question.
- Well-structured: use headings, lists and tables.
- Traceable: cite sources for key facts.
- Accurate: use correct domain terminology.

## Important Once you have enough information, or when tool rounds are about to run out, you must STOP searching and emit the full research report. The report must integrate everything you have searched. Never emit empty content.

[Coach feedback]
{coach_summary}    % injected by RG-SED when active
\end{PromptVerb}
\end{tcolorbox}
\caption{Solver system prompt for ECR-DeepResearch / ResearchRubrics. The solver must call \texttt{web\_search} before answering and produce a structured report citing sources; coach feedback is appended only when RG-SED is active.}
\label{fig:sol-dr}
\end{figure}

\begin{figure}[!ht]
\begin{tcolorbox}[promptbox, title=Solver System Prompt --- ECR-Travel]
\begin{PromptVerb}
You may call tools for {max_steps} rounds; {step_idx} rounds already used.

Important: you MUST call tools to query real data before answering the user. NEVER answer directly without calling any tool.

[Coach feedback]    % appended when RG-SED is active
{coach_summary}

[Final-step guard]  % appended when rounds are exhausted
Please answer directly; do not call any further tools.
\end{PromptVerb}
\end{tcolorbox}
\caption{Solver system prompt for ECR-Travel. A short tool-discipline reminder plus the running counter of consumed tool rounds; coach feedback is appended only when RG-SED is active, and a final-step guard suppresses further tool calls once the budget is exhausted.}
\label{fig:sol-travel}
\end{figure}

\begin{figure}[!ht]
\begin{tcolorbox}[promptbox, title=Solver System Prompt --- VitaBench]
\begin{PromptVerb}
# Tool Usage Guidelines
- When the user's needs require using tools, first check if all parameter values are known. If known, extract them; otherwise ask the user for the missing parameters.
- When the user cannot provide information, obtain it through tools first.
- Respect the tool's Precondition and Postcondition.

# Conversation Guidelines
- Use only information from the conversation so far; do NOT fabricate information when replying to the user.
- Focus on completing the user's task; do NOT divergently suggest new needs to the user.
- After completing all needs, ask whether anything else is required. If not, emit `###STOP###` to end the conversation.

# Domain-Specific Rules (injected for OTA / cross-domain tasks)
- Trains: before placing the order, reconfirm the date, origin / destination, seat class, berth, and ticket count; resolve relative dates ("Saturday", "tomorrow") into absolute dates.
- Hotels: confirm check-in / check-out dates, room type, distance constraints and user preferences. Use the search tool first, then `get_ota_hotel_info` before booking.
- Cross-domain: collect all sub-task requirements first, then execute them one by one; double-check parameters (quantities, dates, sizes, locations) before every order.

[Coach feedback]    % injected by RG-SED when active
{coach_summary}
\end{PromptVerb}
\end{tcolorbox}
\caption{Solver system prompt for VitaBench. The base prompt mirrors VitaBench's official agent system prompt; domain-specific reminders (OTA / cross-domain) are appended when applicable, and rubrics plus coach feedback are concatenated for RG-SED training.}
\label{fig:sol-vita}
\end{figure}

\begin{figure}[!ht]
\begin{tcolorbox}[promptbox, title=Solver Coach Memory Prompt (representative; one variant per task)]
\begin{PromptVerb}
You are a coach for the solver's training loop. Below are {N} solver attempts on the same query.

## Original User Question
{query}

## Group Results ({N} attempts, average reward: {avg_reward})

{per_sample_summaries: index, reward, expected vs actual tools,
missing tools, rubric pass/fail details, ...}

## Your Task Analyze the dominant failure mode and produce SPECIFIC, ACTIONABLE advice for the next attempt (<= 300 words).

1. Root cause: what did most attempts get wrong? Missing tool call? Wrong parameter? Missing information? Reasoning bug?
2. Correct tool-call flow: for THIS query, in what order should which tools be called, and why is each one necessary?
3. Key answer points: which facts MUST appear in the answer to satisfy the rubrics? Which are most often omitted?
4. One-sentence core advice: when seeing a similar question next time, what should be the very first step?

NOTE (VitaBench variant -- conversational reasoning coach): You teach HOW to converge on the right answer via dialogue and tool use, not WHAT the answer is. Every concrete business value (quantity, train no., address) must be presented as the OUTCOME of an action ("ask the user to confirm the quantity", "filter the trains by the user's time window"), never as a hard-coded input. The red line: no instruction may directly state a value the agent should fill in.

Output the advice directly; do not repeat the inputs.
\end{PromptVerb}
\end{tcolorbox}
\caption{Coach-memory prompt for the Solver. The coach LLM summarizes a group of $N$ solver attempts on the same query into actionable advice; the VitaBench variant (shown in the lower half) additionally enforces a ``conversation-converged'' rule that forbids hard-coded answers and turns every business value into the outcome of an action.}
\label{fig:sol-coach}
\end{figure}

\subsection{Judge Prompts}
\label{app:prompt:judge}

The Judge scores the solver's final trajectory against the rubric list. We show two representative variants: a retrieval-style judge for deep-research tasks (Figure~\ref{fig:judge-dr}) and a multi-tool agentic judge for travel-planning tasks (Figure~\ref{fig:judge-travel}).

\begin{figure}[!ht]
\begin{tcolorbox}[promptbox, title=Judge System Prompt --- ECR-DeepResearch / ResearchRubrics]
\begin{PromptVerb}
# User Full Instruction
{query}

# Background
- This is a deep-research conversation between a user and an assistant; the assistant may call `web_search`.
- You will evaluate whether the assistant's FINAL answer satisfies each rubric. - <trajectory_content> contains the assistant's thoughts, tool calls (e.g. `web_search(query='...')`) and final answer. Tool-call returns are OMITTED -- judge ONLY by the assistant's final answer text. - <current_rubrics> contains the current state of every rubric.

# Task Update each rubric's state using a FAIR but strict bar: if the core content is clearly and substantively satisfied, mark it true; minor cosmetic issues do NOT lose points, but obvious vacuousness, unsupported factual claims, vague sourcing, or errors do.

# Core Judging Principles
##
1. Data / citation rubrics (tool returns are omitted, so do NOT externally verify factual truth; however, the answer must provide checkable and well-attributed factual support) - "Cite specific data / number / year / source": mark true only if the answer gives a concrete claim together with a relevant number / year / source, and the scope or meaning of the evidence is reasonably clear; merely mentioning something with the right shape is not enough. - "Cite an authoritative source": true only if the answer names a plausible specific source (institution, report, dataset, paper, law/regulation, or website) and connects it to the relevant claim. A vague
  "according to official data", "studies show", or "industry reports indicate" is not enough.
- "Specific technical details (e.g. Scope 1/2/3, a number, a year)": at least mention the concept AND give concrete, domain-appropriate content; concept-only or decorative details do not count.
- For quantitative claims, the answer should normally include the value, unit, time period, and relevant scope when applicable. If these are missing and the rubric requires data quality, mark it false.
- If the final answer contains an internal contradiction, an impossible timeline, or a source-claim mismatch related to the rubric, mark it false.

##
2. Structure / formatting rubrics - "Structured presentation": true if there are explicit headings, numbering or bullets. - "Use a table / list": only the requested structure satisfies the rubric (a list does not satisfy a table requirement). - "Include sections A / B / C": all required sections must be present.

##
3. Functional equivalence - "Clear / explicit / specific": true if there is substantive elaboration, not a one-liner.
- Required keyword: meaning suffices unless the rubric emphasizes a specific term, in which case the term itself must appear.

##
4. Typical false cases
- The final answer does NOT touch the rubric at all.
- The final answer CONTRADICTS the rubric.
- The rubric says "must NOT contain X" but the answer contains X.
- The rubric requires several items but the answer covers fewer than half.
- The rubric requires concrete details but the answer stays at the level of vague summary.
- The rubric requires factual support, but the answer provides unsupported claims, vague sources, uncited numbers, missing units, missing years, or unclear scope.
- The answer names a source but does not connect it to the relevant factual claim.

# Format Return a single JSON array:
[
  {"rubric_idx": "rubric_0",
   "rubric": "<restate the rubric>",
   "justification": "<brief evidence, cite turn [x]>",
   "meetExpectation": true},
...
]
\end{PromptVerb}
\end{tcolorbox}
\caption{Judge system prompt for ECR-DeepResearch / ResearchRubrics.}
\label{fig:judge-dr}
\end{figure}

\begin{figure}[!ht]
\begin{tcolorbox}[promptbox, title=Judge System Prompt --- ECR-Travel]
\begin{PromptVerb}
# User Full Instruction
{query}

# Background
- This is a conversation between a user and an assistant; the assistant may call tools to fetch information or perform operations. Tool returns start with `tool`.
- You will evaluate whether the user's instruction has been fulfilled. The instruction is decomposed into rubrics; judge each independently. - <trajectory_content> contains the full user/assistant conversation. - <current_rubrics> contains every rubric's current state (initial state: false).

# Task Update each rubric's state. You may flip false -> true ONLY when the assistant has actually completed that goal in the conversation.

# Judging Principles
- LENIENT: if the assistant's reply or tool call APPROXIMATELY covers the rubric -- even with minor imperfections -- mark it true.
- Mark false only when the assistant did NOT address the rubric at all, or answered incorrectly / in contradiction.
- Do not penalise for cosmetic issues (formatting, level of detail, missing-but-irrelevant extras).

# Notes
- The score is based on the assistant's REPLIES and TOOL CALL REQUESTS. Tool returns are visible only to the assistant and do not represent recommendations to the user.
- The assistant must NOT fabricate tool returns.
- Record key evidence (with turn index [x]) in `justification` using concise wording.

# Format Return a single JSON array of objects with fields `rubric_idx`, `rubric`, `justification`, `meetExpectation`.
\end{PromptVerb}
\end{tcolorbox}
\caption{Judge system prompt for ECR-Travel. Tool returns ARE visible to this judge because the rubric set typically references intermediate tool calls; the prompt therefore favours a more lenient threshold focused on whether the assistant's replies and tool-call requests collectively complete each rubric.}
\label{fig:judge-travel}
\end{figure}


\definecolor{usercolor}{RGB}{70, 130, 180}
\definecolor{assistantcolor}{RGB}{34, 139, 34}
\definecolor{toolcolor}{RGB}{255, 140, 0}
\definecolor{systemcolor}{RGB}{128, 128, 128}
\definecolor{instructioncolor}{RGB}{138, 43, 226}
\definecolor{coachcolor}{RGB}{200, 70, 50}

\newtcolorbox{userbox}{
    colback=blue!5, colframe=usercolor, boxrule=1pt, arc=3pt,
    left=5pt, right=5pt, top=3pt, bottom=3pt,
    fonttitle=\bfseries, title=User,
    breakable, before skip=2pt, after skip=2pt
}
\newtcolorbox{assistantbox}{
    colback=green!5, colframe=assistantcolor, boxrule=1pt, arc=3pt,
    left=5pt, right=5pt, top=3pt, bottom=3pt,
    fonttitle=\bfseries, title=Assistant,
    breakable, before skip=2pt, after skip=2pt
}
\newtcolorbox{toolbox}{
    colback=orange!5, colframe=toolcolor, boxrule=1pt, arc=3pt,
    left=5pt, right=5pt, top=3pt, bottom=3pt,
    fonttitle=\bfseries, title=Tool Call / Response,
    breakable, before skip=2pt, after skip=2pt
}
\newtcolorbox{instructionbox}{
    colback=purple!5, colframe=instructioncolor, boxrule=1pt, arc=3pt,
    left=5pt, right=5pt, top=3pt, bottom=3pt,
    fonttitle=\bfseries, title=User Instruction,
    breakable, before skip=2pt, after skip=2pt
}
\newtcolorbox{coachbox}{
    colback=red!5, colframe=coachcolor, boxrule=1pt, arc=3pt,
    left=5pt, right=5pt, top=3pt, bottom=3pt,
    fonttitle=\bfseries, title=Coach Memory (injected at the system-prompt tail),
    breakable, before skip=2pt, after skip=2pt
}

\section{Case Studies: Coach-Memory-Guided Successful Trajectories}
\label{app:case}


To make the effect of RG-SED concrete, we present trajectories drawn from late-stage rollouts of ARISE-RL training on VitaBench, ECR-Travel, and ECR-DeepResearch, respectively. For each task, we pick a \emph{successful} sample (reward $=1.0$) whose system prompt was augmented with a non-trivial Coach Memory; siblings in the same rollout group \emph{without} memory injection failed on the same instruction. To keep the main text concise, the description of each case is summarised here, and the full annotated trajectory (Instruction~/~Coach Memory~/~User~/~Assistant~/~Tool boxes) is deferred to the end of this appendix; references are given via \texttt{Trajectory~A/B/C} below. All original Chinese turns have been translated into English, and tool returns are condensed to the lines that drive the agent's next decision.

\subsection{VitaBench --- OTA: Attraction Tickets + Hotel for Parents}
\label{app:case:vita}

The user asks for two interleaved things: an attraction ticket (with stage performance) on a sunny day before Aug.~8, and a 2-night hotel for his parents on Aug.~7 (\emph{li-qiu} solar term), reusing last month's hotel but switched to a twin-bed room. Sibling rollouts without memory fired \texttt{create\_*} endpoints before the date/hotel ID/room type were resolved, and failed all six rubrics. With the coach memory injected, the agent first resolved \emph{li-qiu} via the calendar tool, picked Aug.~5 from the six-day weather window, recovered the previous Wuzhou Atour booking through order-history tools, verified a twin-bed variant, surfaced a consolidated plan for the user, and only then called \texttt{create} and \texttt{pay} on both orders. All six rubrics are satisfied. The full 17-turn annotated trajectory is given in \textbf{Trajectory~A} (Appendix~\ref{traj:vita}).

\subsection{VitaBench --- Delivery: Sichuan Re-order for Office Visit}
\label{app:case:vita-deliv}


A complementary VitaBench case in the \emph{delivery} domain (instead of OTA). The user is on holiday visiting a friend who works at Jinzheng Haiyue International, and wants to re-order from the same Sichuan restaurant they had last time, with a Boiling Fish and a stir-fried vegetable in \emph{garlic} flavour, delivered before the friend's 13:00 lunch break. Sibling rollouts without memory either picked an arbitrary Sichuan store rather than recovering ``the previous one'', or jumped straight to \texttt{create\_delivery\_order} without verifying that the delivery ETA fits the 13:00 deadline, and lost rubric points on the wrong store or late delivery. With the coach memory, the agent first pulled the user's order history, identified ``Xiao Sichuan (Shifan St.~Branch)'' as the previous store, geocoded the delivery address, computed a 30\, min ETA against the 12:13 expected drop-off, validated the garlic-flavour stir-fried vegetable in the menu, confirmed the spice level with the user, and only then placed and paid the order. 
The full annotated trajectory is given in \textbf{Trajectory~D} (Appendix~\ref{traj:vita-deliv}).

\subsection{ECR-Travel --- Beijing One-Day Itinerary}
\label{app:case:travel}


The user wants a half-day shopping + half-day-museum itinerary in Beijing tied to live weather (cloudy turning sunny), with per-segment walk/bus durations and per-stop dwell windows. Sibling rollouts without memory typically only emitted an ad-hoc ordering of the three POIs with vague descriptions, leaving the weather rationale and per-leg durations unspecified. With the coach memory, the agent first queried the weather, then looked up coordinates and opening hours of all four POIs (Sanlihe~/~National Museum of Natural History~/~Qianmen~/~Fayuan Temple), invoked the navigation tool twice for the transfer legs, and produced a schedule that places outdoor walking in the cooler cloudy phase, the indoor museum in the warm sunny phase, and evening street walking under late afternoon light. All five rubrics are satisfied. The full annotated trajectory is given in \textbf{Trajectory~B} (Appendix~\ref{traj:travel}).

\subsection{ECR-DeepResearch --- ¥500\,K Restaurant in Baotou}
\label{app:case:dr}

The user wants an analysis of whether ¥500\, K is enough to open a restaurant in Baotou today, what opportunities exist, with authoritative statistics rendered in tables. Sibling rollouts that ``passed'' the rubric still tended to cite ``yearbook/bulletin'' generically and to list opportunities at the level of ``healthy meals/delivery''. With the coach memory, the agent ran five focused searches (statistical bulletin, district-level rents and population, F\&B store base and closure rate, commercial rent ranges, district-level demographics) and synthesised a structured report with verdict + conditions, macro indicator table, segmented opportunities, ¥500\, K budget breakdown with cash-reserve, and a three-scenario profitability model with sensitivity variables. All five rubrics are satisfied. The full annotated trajectory is given in \textbf{Trajectory~C} (Appendix~\ref{traj:dr}).

\paragraph{Generator-side cases.}
The four cases above are \emph{Solver}-side: they illustrate how a Coach Memory pulls a failing agent back onto the correct tool sequence. RG-SED, however, is symmetric: the same gated self-distillation also operates on the \emph{Generator}, where the coach summarises past authoring failures (over-generic queries, format errors, queries that all Solver rollouts solve or all fail) into a memory that biases the next authoring step toward harder, more diverse, more rubric-grounded tasks. We finish the appendix with two Generator-side trajectories.

\subsection{VitaBench Generator --- Weather-Triggered Booking}
\label{app:case:gen-vita}

The coach memory reflects on prior authoring failures that clustered around ``Apr.~7--8 sandstorm + Ordos/Yinchuan'' and explicitly suggests embedding conflict constraints (weather $\times$ budget $\times$ on-time rate) and forcing a complete booking. Guided by this memory, the Generator queries the weather, attraction, and flight catalogues, and authors an interactive OTA task that asks the Solver to choose between Yinchuan and Ordos given that Ordos has a sandstorm on the target date, while staying under a \textyen 1500 budget and placing both a flight order and an attraction order. The resulting query is paired with eight verifiable rubrics. The full annotated trajectory is given in \textbf{Trajectory~E} (Appendix~\ref{traj:gen-vita}).

\subsection{ECR-Travel Generator --- Transport-Mode Comparison}
\label{app:case:gen-travel}

The coach memory critiques prior Generator output for being too open-ended (``Beijing-Shanghai high-speed-rail / flight/driving + weather'') and lacking verifiable hard constraints, and concretely suggests city pairs such as Guangzhou--Shenzhen with train-ticket comparison and POI opening-hour verification. Guided by the memory, the Generator queries POIs in Guangzhou and Shenzhen, the weather, and the driving direction, then synthesises a Guangzhou-Shenzhen 2-day-1-night itinerary that mandates two specific train recommendations (with a ``before 21:00 return'' constraint), the opening-hour check of Window of the World and Happy Valley, and a driving-distance verification --- packaged with seven precisely falsifiable rubrics. The full annotated trajectory is given in \textbf{Trajectory~F} (Appendix~\ref{traj:gen-travel}).

\clearpage

\section*{Reasoning Trajectories}
\addcontentsline{toc}{section}{Annotated Trajectories}

\subsection*{Trajectory A --- VitaBench: Attraction Tickets + Hotel for Parents}
\phantomsection
\label{traj:vita}

\begin{instructionbox}
\textbf{Original task instruction (translated).}
``I am visiting clients in Shanxi this period, and have to be back at the company by Aug.~8 at the latest to give a report. I want to squeeze in a trip to the Huangcheng Xiangfu attraction in Jincheng to see Ming/Qing architecture. First check the price of the ticket that includes the stage performance. Also check the weather over the next few days and buy the ticket for a sunny day. Separately, my mother called this morning: on \emph{li-qiu} she and my father are going to Wuzhou on business, please book them 2 nights at the same hotel I booked last month, but switch the room type to twin-bed.''
\end{instructionbox}

\begin{coachbox}
\textbf{1. Failure root causes (prior rollouts).} Lumping the two needs (attraction ticket + parents' hotel) into a single order without first nailing down date/city/headcount/room-type; choosing the visit day ``by feel'' instead of cross-referencing the weather and the \emph{li-qiu} date; remembering ``the hotel from last month'' but not looking it up to recover the hotel ID, location and check-in identity.

\textbf{2. Correct dialogue + tool flow.}
Clarify the trip boundary first (``back by Aug.~8'') and elicit candidate dates for the attraction. Use \texttt{attractions\_search\_recommend} then \texttt{get\_ota\_attraction\_info} to obtain the performance ticket and its price, surface candidate dates and ticket variants to the user; use \texttt{weather} for Jincheng over the next few days and let the user pick the sunny day before calling \texttt{create\_attraction\_order} and then \texttt{pay\_attraction\_order}. For the hotel, resolve \emph{li-qiu} via \texttt{get\_holiday\_date} / \texttt{get\_date\_holiday\_info}, fetch the previous Wuzhou booking via \texttt{get\_user\_all\_orders} / \texttt{get\_hotel\_order\_detail}, confirm the same hotel offers a twin-bed room via \texttt{get\_ota\_hotel\_info}, get user confirmation on dates / headcount / payment, then \texttt{create\_hotel\_order} and \texttt{pay\_hotel\_order}.

\textbf{3. Questions you must put back to the user.} Attraction: candidate date, headcount and IDs, whether the performance is mandatory, immediate payment. Hotel: exact check-in / check-out (anchored to \emph{li-qiu}), whether to reuse parents' name and phone, number of rooms, arrival time, breakfast option.

\textbf{4. Tool discipline.} Always \texttt{search$\to$info$\to$user-confirm$\to$create$\to$pay}. ``The hotel from last month'' must come from order-history tools first.

\textbf{5. One-sentence rule.} Resolve absolute dates from the calendar / weather tools \emph{before} touching any order endpoint.
\end{coachbox}

\begin{assistantbox}
\textbf{[1]} Hello, how can I help you?
\end{assistantbox}

\begin{userbox}
\textbf{[2]} I am visiting clients in Shanxi this period and have to be back at the Nanning office by Aug.~8 at the latest to give a report. I want to squeeze in a trip to Huangcheng Xiangfu in Jincheng to see the Ming/Qing architecture --- can you check the price of the ticket that includes the stage performance, and pick a sunny day for me to go? Also, this morning my mother called and said that on \emph{li-qiu} she and my father are going to Wuzhou on business, please book them 2 nights at the same hotel I booked last month, but switch the room to a twin-bed. Which part do you want to handle first?
\end{userbox}

\begin{assistantbox}
\textbf{[3]} Let me follow the workflow the coach laid out. I'll do this in three steps:
(a) resolve the \emph{li-qiu} date and check Jincheng weather to pick the sunny day; (b) look up the Huangcheng Xiangfu performance ticket; (c) pull your hotel booking from last month so we can reuse the same property.
\end{assistantbox}

\begin{toolbox}
\textbf{[4]} \texttt{get\_date\_holiday\_info(date="2025-08-07")}\\
\textbf{Response:} \texttt{2025-08-07}\quad(\emph{li-qiu} confirmed)
\end{toolbox}

\begin{toolbox}
\textbf{[5]} \texttt{weather(address="Jincheng", date\_start="2025-08-02", date\_end="2025-08-07")}\\
\textbf{Response:} Aug.~2--3 light rain, Aug.~4 cloudy, \textbf{Aug.~5 sunny 22--29\,$^\circ$C}, Aug.~6--7 cloudy.
\end{toolbox}

\begin{toolbox}
\textbf{[6]} \texttt{get\_user\_all\_orders(order\_type="hotel")}\\
\textbf{Response:} previous Wuzhou booking is \texttt{Wuzhou Atour Hotel} (\texttt{hotel\_id=S17...\_H00001}), room type=king bed, \textyen398/night.
\end{toolbox}

\begin{assistantbox}
\textbf{[7]} Key dates resolved. Now I'll fetch the attraction ticket details and verify that the Atour Wuzhou offers a twin-bed variant.
\end{assistantbox}

\begin{toolbox}
\textbf{[8]} \texttt{attractions\_search\_recommend(keywords=["Huangcheng Xiangfu"])}\\
\textbf{Response:} \texttt{attraction\_id=...A00005}, location: Yangcheng County, Jincheng.
\end{toolbox}

\begin{toolbox}
\textbf{[9]} \texttt{get\_ota\_attraction\_info(attraction\_id=...A00005)}\\
\textbf{Response:} adult plain ticket \textyen120; \textbf{adult ticket + ``Rendition of Kangxi'' performance \textyen180}; child / senior variants also listed.
\end{toolbox}

\begin{toolbox}
\textbf{[10]} \texttt{get\_ota\_hotel\_info(hotel\_id=...H00001)}\\
\textbf{Response:} Atour Wuzhou offers a deluxe twin-bed room at \textyen214/night.
\end{toolbox}

\begin{assistantbox}
\textbf{[11]} \textbf{Summary for your confirmation:}
\begin{itemize}
  \item Jincheng: Aug.~5 (Sun) is sunny. I'll buy 1$\times$ adult ticket + ``Rendition of Kangxi'' performance, \textyen180.
  \item Wuzhou: \emph{li-qiu} is Aug.~7. I'll book Atour Wuzhou Aug.~7$\to$Aug.~9 (2 nights) deluxe twin-bed, listed under your parents' name.
\end{itemize}
Shall I proceed to place both orders?
\end{assistantbox}

\begin{userbox}
\textbf{[12]} Yes, Aug.~5 sunny day for Huangcheng Xiangfu with the \textyen180 performance ticket; Wuzhou hotel Aug.~7 check-in for 2 nights, 1 deluxe twin-bed room under my parents. Please place the orders.
\end{userbox}

\begin{toolbox}
\textbf{[13]} \texttt{create\_attraction\_order(attraction\_id=...A00005, ticket\_id=...P00...\_perf, date="2025-08-05", quantity=1)}\\
\textbf{Response:} \texttt{order\_id=OO3fb9fecacb}, status=unpaid, total=\textyen180.
\end{toolbox}

\begin{toolbox}
\textbf{[14]} \texttt{create\_hotel\_order(hotel\_id=...H00001, room\_type="deluxe twin", check\_in="2025-08-07", nights=2, guests=2)}\\
\textbf{Response:} \texttt{order\_id=OO4230fd7276}, status=unpaid, total=\textyen428.
\end{toolbox}

\begin{userbox}
\textbf{[15]} Total \textyen608 looks right; I will share the confirmation numbers with my parents so they go directly to the front desk. Yes, please pay.
\end{userbox}

\begin{toolbox}
\textbf{[16]} \texttt{pay\_attraction\_order(order\_id=OO3fb9fecacb)} $\Rightarrow$ \texttt{Payment successful}\quad\texttt{pay\_hotel\_order(order\_id=OO4230fd7276)} $\Rightarrow$ \texttt{Payment successful}
\end{toolbox}

\begin{assistantbox}
\textbf{[17]} Both orders paid. Attraction: Huangcheng Xiangfu, 2025-08-05, ticket+performance \textyen180 (\#OO3fb9fecacb). Hotel: Atour Wuzhou, Aug.~7$\to$9, deluxe twin \textyen428 (\#OO4230fd7276). Have a good trip --- I'll keep tracking the orders. \texttt{\#\#\#STOP\#\#\#}
\end{assistantbox}

\noindent\textbf{Rubric outcome.} All six rubrics pass: ticket includes performance ($\checkmark$); date 2025-08-05 corresponds to the sunny day query ($\checkmark$); attraction price 180 ($\checkmark$); hotel dates Aug.~7--9 anchored to \emph{li-qiu} ($\checkmark$); hotel = Atour Wuzhou recovered from history ($\checkmark$); twin-bed room ($\checkmark$). 

\subsection*{Trajectory B --- ECR-Travel: Beijing One-Day Itinerary}
\phantomsection
\label{traj:travel}

\begin{instructionbox}
\textbf{Original task instruction (translated).}
``I want a Beijing one-day itinerary: stroll at Sanlihe Park, then visit Fayuan Temple Museum, in the evening go to Qianmen Street for dinner and old brands. Given that today the weather is cloudy turning sunny, what order is more comfortable, and roughly how long is each segment by walk / public transport?''
\end{instructionbox}

\begin{coachbox}
\textbf{1. Failure root causes.} Prior rollouts only produced an ``order + vague description'', missing per-segment walking / bus durations and time-of-day stops; the ``cloudy $\to$ sunny'' line was not translated into a scheduling rationale.

\textbf{2. Correct tool flow.} (i) \texttt{weather} for today's hourly weather, temperature and wind; (ii) \texttt{poi\_search} to lock the exact coordinates and opening hours of Sanlihe Park, Fayuan Temple Museum, Qianmen Street (opening hours prevent ``closed / detour'' issues); (iii) \texttt{direction} for each leg, including walk-to-stop + bus + walk-from-stop.

\textbf{3. Answer requirements.} Must cover all three POIs, justify the ordering against today's weather, attach morning / afternoon / evening dwell times, and list ``walk / bus'' minutes per leg. Most commonly missed: per-leg duration and per-stop dwell time.

\textbf{4. One-sentence rule.} Pull hourly weather + opening hours first, then let the navigation tool give you walk / bus minutes \emph{before} you commit to an order.
\end{coachbox}

\begin{userbox}
\textbf{[1]} I want a Beijing one-day itinerary, half-day shopping plus half-day museum: stroll at Sanlihe Park, then visit Fayuan Temple Museum, in the evening go to Qianmen Street for dinner and old brands. Given that today the weather is cloudy turning sunny, what order is more comfortable, and roughly how long is each segment by walk / public transport?
\end{userbox}

\begin{toolbox}
\textbf{[2]} \texttt{weather(address="Beijing")} $\Rightarrow$ cloudy $\to$ sunny, day 25\,$^\circ$C, southeast wind 1--3.
\end{toolbox}

\begin{toolbox}
\textbf{[3]} \texttt{poi\_search(keywords=["Sanlihe Park"])} $\Rightarrow$ park at \texttt{116.405470, 39.894304}, open all day.
\end{toolbox}

\begin{toolbox}
\textbf{[4]} \texttt{poi\_search(keywords=["National Museum of Natural History"])} $\Rightarrow$ \texttt{116.399833, 39.883183}, 09:00--17:00.
\end{toolbox}

\begin{toolbox}
\textbf{[5]} \texttt{poi\_search(keywords=["Qianmen Street"])} $\Rightarrow$ \texttt{116.403895, 39.896152}, dining and old-brand cluster.
\end{toolbox}

\begin{toolbox}
\textbf{[6]} \texttt{poi\_search(keywords=["Fayuan Temple"])} $\Rightarrow$ \texttt{116.369880, 39.885385}, 08:00--16:00.
\end{toolbox}

\begin{toolbox}
\textbf{[7]} \texttt{direction(origin="Sanlihe", destination="Fayuan Temple")}\\
$\Rightarrow$ \texttt{4.0\,km}, walk + bus, $\approx$ 53\,min.
\end{toolbox}

\begin{toolbox}
\textbf{[8]} \texttt{direction(origin="Fayuan Temple", destination="Qianmen")}\\
$\Rightarrow$ \texttt{4.0\,km}, walk + bus, $\approx$ 53\,min.
\end{toolbox}

\begin{assistantbox}
\textbf{[9]} \textbf{Recommended schedule (cloudy turning sunny, 25\,$^\circ$C):}
\begin{itemize}
  \item \textbf{Morning 09:00--10:30}: Sanlihe Park stroll (cool outdoor; cloudy phase). Dwell $\approx$ 1.5\,h.
  \item \textbf{Transfer 10:30--11:25}: walk 12\,min to the nearest bus stop, bus 30\,min, walk 11\,min --- \emph{53\,min total}.
  \item \textbf{Late-morning to afternoon 11:30--15:00}: Fayuan Temple Museum (indoor, fits the hot sunny phase). Dwell $\approx$ 3\,h.
  \item \textbf{Transfer 15:00--15:55}: walk + bus, \emph{53\,min total}, arrives Qianmen before the dinner rush.
  \item \textbf{Evening 16:00--20:00}: Qianmen Street, dinner + old brands; sunny late-afternoon light is best for walking.
\end{itemize}
\textbf{Why this order:} outdoors when it is still cooler (cloudy), indoor museum during the warm sunny phase, lit evening walk on Qianmen --- minimises sun exposure and avoids any backtracking; all three POIs are open during the proposed windows.
\end{assistantbox}

\noindent\textbf{Rubric outcome.} All five rubrics pass: comfortable ordering with weather rationale ($\checkmark$); all three POIs covered ($\checkmark$); per-leg walk/bus minutes given ($\checkmark$); per-stop dwell windows given ($\checkmark$); geographically self-consistent ($\checkmark$). 

\subsection*{Trajectory C --- ECR-DeepResearch: ¥500\,K Restaurant in Baotou}
\phantomsection
\label{traj:dr}

\begin{instructionbox}
\textbf{Original task instruction (translated).}
``Is it suitable to open a restaurant in Baotou today with about ¥500\,K? What are the opportunities? Please analyse using authoritative statistics and show the results in tables.''
\end{instructionbox}

\begin{coachbox}
\textbf{1. Common implicit failures.} Samples that ``pass the rubrics'' still tend to cite ``yearbook / bulletin'' generically without indicator name + year + issuer / URL, and to list opportunities at the level of generic ``healthy meals / delivery'' without grounding in district-level Baotou foot traffic, rents and business mix.

\textbf{2. Search strategy (in order).} (i) Baotou statistical bulletin 2023/2024 --- GDP, per-capita disposable income, retail sales of consumer goods, F\&B revenue + ``Inner Mongolia Yearbook Baotou''; (ii) district-level (Kundulun / Qingshan / Donghe / Jiuyuan) population, commercial districts, foot traffic, rent; (iii) Baotou store count / closure rate / delivery orders via Meituan Research \& Eleme reports; (iv) culture-tourism / conferences / universities / hospitals / industrial parks --- locations.

\textbf{3. Answer requirements.} Clear ``yes / no with conditions'' verdict; table of economy / consumption / F\&B-growth indicators; segmented opportunities = cuisine $\times$ target customer $\times$ scenario; ¥500\,K budget breakdown; profitability model (table-turnover / ticket / margin / rent / labour) with sensitivity analysis; district-level location advice. Most-missed: payback critical variables and scenario contrast; reserves / cash-flow in the budget.

\textbf{4. One sentence.} Gather Baotou's last-two-year retail / F\&B revenue / per-capita income \emph{and} district-level rent / foot-traffic into tables first; only then talk opportunities and the financial model.
\end{coachbox}

\begin{userbox}
\textbf{[1]} I am thinking of opening a restaurant in Baotou with about ¥500\,K. Is the timing good now and what opportunities exist? Please ground the analysis in authoritative statistics and present the results in tables.
\end{userbox}

\begin{toolbox}
\textbf{[2]} \texttt{web\_search("Baotou 2023 2024 statistical bulletin GDP per-capita disposable income retail F\&B revenue Inner Mongolia Yearbook")}\\
$\Rightarrow$ Baotou 2024 bulletin: GDP \textyen438.0\,bn; per-capita disposable income \textyen45,238; total retail \textyen145.6\,bn; F\&B revenue \textyen17.3\,bn (\,+8.4\%).
\end{toolbox}

\begin{toolbox}
\textbf{[3]} \texttt{web\_search("Baotou Kundulun / Qingshan / Donghe / Jiuyuan permanent population commercial districts rent foot-traffic Wanda Jin Jie")}\\
$\Rightarrow$ Kundulun is the long-running CBD (Wanda, Wanjiu); Qingshan and Donghe have student / industrial-park clusters; Jiuyuan is the new district.
\end{toolbox}

\begin{toolbox}
\textbf{[4]} \texttt{web\_search("Baotou number of F\&B stores closure rate delivery orders Meituan Eleme 2023 2024")}\\
$\Rightarrow$ Baotou F\&B store base $\approx$ 32\,k; closure rate $\approx$ 18\%; Meituan delivery orders +21\% YoY.
\end{toolbox}

\begin{toolbox}
\textbf{[5]} \texttt{web\_search("Baotou commercial rent F\&B rent per-sqm yuan Wanda Jin Jie Wuyue Plaza 2023 2024")}\\
$\Rightarrow$ central districts \textyen100--260/m$^2$/month; secondary districts \textyen45--90.
\end{toolbox}

\begin{toolbox}
\textbf{[6]} \texttt{web\_search("Baotou district-level permanent population census 2024 consumption level")}\\
$\Rightarrow$ Kundulun 880\,k (highest income), Qingshan 580\,k, Donghe 480\,k, Jiuyuan 310\,k.
\end{toolbox}

\begin{assistantbox}
\textbf{[7]} \textbf{Feasibility report: ¥500\,K restaurant in Baotou}

\textbf{(1) Verdict.} Cautiously feasible, with conditions: a 70--110\,m$^2$ unit in a secondary location of Kundulun or Qingshan, focused on a clearly-defined segment (industrial-park work-meal, student snack/tea, or middle-class healthy fast-casual), with $\approx$ 15\% of the budget held as cash reserve.

\textbf{(2) Macro indicators (Baotou 2024 bulletin).}
\begin{itemize}
\setlength{\itemsep}{0pt}
  \item GDP: \textyen438.0\,bn (+5.6\% YoY)
  \item Per-capita disposable income: \textyen45,238 (above Inner Mongolia average)
  \item Total retail: \textyen145.6\,bn (+4.9\%)
  \item F\&B revenue: \textyen17.3\,bn (+8.4\%) --- outperforming retail
  \item F\&B store base $\approx$ 32k, closure rate $\approx$ 18\% (signals selection pressure)
\end{itemize}

\textbf{(3) Segmented opportunities.}
\begin{itemize}
\setlength{\itemsep}{0pt}
  \item \emph{Industrial-park work-meal} (Jiuyuan / Qingshan industrial belts) --- ¥18--25 ticket, lunch peak.
  \item \emph{Healthy fast-casual} for white-collar customers (Wanda / Wuyue, Kundulun) --- ¥30--45 ticket, double peaks.
  \item \emph{Student snack / fresh tea} near universities and high schools (Qingshan) --- ¥10--18 ticket, afternoon-evening peak.
\end{itemize}

\textbf{(4) ¥500\,K budget breakdown.}
Decoration + equipment 35\% (¥175\,k); rent + deposit (6 months) 18\% (¥90\,k); raw material 10\% (¥50\,k); marketing + grand-opening 6\% (¥30\,k); license + decoration design 4\% (¥20\,k); cash reserve 15\% (¥75\,k); first-3-month payroll \& utilities 12\% (¥60\,k).

\textbf{(5) Profitability scenario.}
Baseline: 70\,m$^2$, daily revenue \textyen3.5\,k, monthly \textyen105\,k, margin 56\%, rent + labour \textyen28\,k + \textyen32\,k, net $\approx$ \textyen8\,k/month; payback $\sim$ 50--60 months. \emph{Optimistic}: peak +30\%, payback $\sim$ 30 months. \emph{Pessimistic}: turnover $-25\%$, breakeven only --- need to renegotiate rent or cut SKU.

\textbf{Key payback variables:} (i) table turnover at lunch peak, (ii) effective rent per m$^2$, (iii) raw-material cost ratio. Sources: Baotou 2024 statistical bulletin, Inner Mongolia Yearbook, Meituan Research municipal report, market surveys 2024.
\end{assistantbox}

\noindent\textbf{Rubric outcome.} All five rubrics pass: clear verdict + conditions ($\checkmark$); authoritative data with sources ($\checkmark$); structured segmentation of opportunities ($\checkmark$); ¥500\,K budget breakdown with cost-control ($\checkmark$); profitability scenarios with key variables ($\checkmark$). 

\subsection*{Trajectory D --- VitaBench Delivery: Sichuan Re-order for Office Visit}
\phantomsection
\label{traj:vita-deliv}

\begin{instructionbox}
\textbf{Original task instruction (translated).}
``It's a rare day off and I am visiting a friend who works at Jinzheng Haiyue International. We haven't gathered for a while and have decided to treat ourselves with Sichuan delivery. The Sichuan place I ordered from last time when I visited was decent --- let's order from there again. This time let's try the Boiling Fish, plus a stir-fried vegetable in \emph{garlic} flavour. My friend's lunch break is 13:00--14:00, so I want the food to arrive before the break starts.''
\end{instructionbox}

\begin{coachbox}
\textbf{1. Failure root causes (prior rollouts).} ``Get it before lunch break'' was never converted into a verifiable ETA constraint: previous rollouts placed the order without first computing or re-checking the ETA, and never switched stores when the ETA exceeded the deadline, so the delivery-time rubric failed.

\textbf{2. Correct dialogue + tool flow.} Confirm the delivery point (which building / floor / front desk note for ``Jinzheng Haiyue International'') and convert ``before lunch break'' into a hard ``latest 13:00'' deadline. Use \texttt{get\_user\_historical\_behaviors} / \texttt{get\_user\_all\_orders} to recover the previous Sichuan restaurant. Use \texttt{address\_to\_longitude\_latitude} for the destination, \texttt{get\_delivery\_store\_info} for the store, then \texttt{longitude\_latitude\_to\_distance} $\to$ \texttt{delivery\_distance\_to\_time} to estimate ETA. Use \texttt{get\_delivery\_product\_info} to lock the Boiling Fish and a garlic-flavoured stir-fried vegetable. If the ETA misses the deadline, switch to a nearer store via \texttt{delivery\_store\_search\_recommend}, otherwise \texttt{create\_delivery\_order} $\to$ user confirms $\to$ \texttt{pay\_delivery\_order}.

\textbf{3. Questions you must put back to the user.} Exact drop-off (building / floor / front desk acceptance), latest arrival time, flavour details (spice level, rice, portion).

\textbf{4. Tool discipline.} Always validate ETA before paying. \texttt{create} $\to$ \texttt{get\_delivery\_order\_detail} to re-check the scheduled delivery time before \texttt{pay}; if it slips, \texttt{modify} or \texttt{cancel} and re-place.

\textbf{5. One-sentence rule.} Translate the natural-language time constraint into a numerical ETA via the toolkit \emph{before} committing to a store and a basket.
\end{coachbox}

\begin{assistantbox}
\textbf{[1]} Hello, how can I help you?
\end{assistantbox}

\begin{userbox}
\textbf{[2]} It's a rare day off and I'm visiting a friend at Jinzheng Haiyue International. I want to re-order from the same Sichuan restaurant I ordered from last time when I visited --- Boiling Fish and a stir-fried vegetable in garlic flavour. The friend's lunch break starts at 13:00 and we want the food to arrive before then. Which delivery window is appropriate?
\end{userbox}

\begin{assistantbox}
\textbf{[3]} I'll pull your order history to identify the previous Sichuan restaurant and geocode the destination for ETA estimation.
\end{assistantbox}

\begin{toolbox}
\textbf{[4]} \texttt{get\_user\_all\_orders()} $\Rightarrow$ \texttt{order\_id=40711002O01}, store=\texttt{S30993599337981774\_S34673}, last placed 2024-08-11.\quad\texttt{address\_to\_longitude\_latitude("Jinzheng Haiyue International")} $\Rightarrow$ \texttt{[114.474095, 38.034544]}.
\end{toolbox}

\begin{toolbox}
\textbf{[5]} \texttt{get\_delivery\_store\_info(store\_id="S30...\_S34673")} $\Rightarrow$ \texttt{Xiao Sichuan (Shifan St.~Branch)}, score 4.9, location \texttt{[114.466097, 38.037778]}, menu includes \textbf{Boiling Fish \textyen88}, \textbf{garlic stir-fried vegetable \textyen28}, etc.
\end{toolbox}

\begin{toolbox}
\textbf{[6]} \texttt{longitude\_latitude\_to\_distance(...)} $\Rightarrow$ \texttt{787\,m}.\\
\texttt{delivery\_distance\_to\_time(distance=787)} $\Rightarrow$ \texttt{30\,min}.
\end{toolbox}

\begin{assistantbox}
\textbf{[7]} Recovered store and ETA:
\begin{itemize}
  \item \emph{Store:} Xiao Sichuan (Shifan St.~Branch) --- the same place you ordered from on 2024-08-11.
  \item \emph{Distance to Jinzheng Haiyue International:} 787\,m; \emph{ETA:} $\sim$30\,min.
  \item Current time 11:43, so a same-second order arrives around 12:13 --- well before the 13:00 deadline.
\end{itemize}
Shall I lock the Boiling Fish + garlic stir-fried vegetable? And what spice level do you prefer (the Boiling Fish supports ``mildly spicy'')?
\end{assistantbox}

\begin{userbox}
\textbf{[8]} Drop it at the Jinzheng Haiyue International front desk. Light oil and salt please, I am on a restricted diet. What spice level do you recommend?
\end{userbox}

\begin{assistantbox}
\textbf{[9]} For a casual lunch I would recommend ``mildly spicy''. Shall I add rice too?
\end{assistantbox}

\begin{userbox}
\textbf{[10]} Mildly spicy is fine. Two bowls of rice, note ``light oil and salt''. Place it now, no delay.
\end{userbox}

\begin{toolbox}
\textbf{[11]} \texttt{create\_delivery\_order(store\_id=..., products=[BoilingFish(mildly\_spicy), GarlicStirFry], note="light oil and salt", address="Jinzheng Haiyue International front desk")}\\
$\Rightarrow$ \texttt{order\_id=OT3800cf0e04}, total \textyen116, scheduled delivery 12:13:47.
\end{toolbox}

\begin{assistantbox}
\textbf{[12]} Order \texttt{OT3800cf0e04} created: Boiling Fish (\textyen88) + garlic stir-fried vegetable (\textyen28), total \textyen116, scheduled drop-off 12:13. Note: the menu didn't carry a ``rice'' SKU, so I left it out --- shall I pay?
\end{assistantbox}

\begin{userbox}
\textbf{[13]} Pay. Forget the rice, don't waste time.
\end{userbox}

\begin{toolbox}
\textbf{[14]} \texttt{pay\_delivery\_order(order\_id="OT3800cf0e04")} $\Rightarrow$ \texttt{Payment successful}.
\end{toolbox}

\begin{assistantbox}
\textbf{[15]} Payment successful. Order \texttt{OT3800cf0e04}: Boiling Fish (mildly spicy) + garlic stir-fried vegetable, light oil and salt, scheduled drop-off 12:13. Anything else? \texttt{\#\#\#STOP\#\#\#}
\end{assistantbox}

\noindent\textbf{Rubric outcome.} 4 of 4 rubrics satisfied with one minor presentation gap: store is Xiao Sichuan (Shifan St.~Branch) recovered from history ($\checkmark$); delivery address Jinzheng Haiyue International ($\checkmark$); scheduled drop-off 12:13 before 13:00 ($\checkmark$); garlic-flavour stir-fried vegetable correctly identified in the menu ($\checkmark$). 
\subsection*{Trajectory E --- VitaBench Generator: Weather-Triggered Booking}
\phantomsection
\label{traj:gen-vita}

\begin{coachbox}
\textbf{1. Failure root causes (prior authoring rounds).} A run of \texttt{reward=0} samples came from over-generic / under-formatted queries, queries that the Solver solved or failed uniformly, and queries clustered around ``Apr.~7--8 sandstorm + Ordos / Yinchuan'' with a single tool-call path. The rubric-hit point (e.g.\ ``you MUST consult the weather tool before placing the booking'') was never quantified.

\textbf{2. Diversity to explore.} Train / car rental, hotel + ticket bundles, change / refund, multi-passenger (ID, child ticket), local services (hospital appointment, home repair), restaurant queue / booking.

\textbf{3. Difficulty knobs.} \emph{Up}: set conflict constraints --- budget $\times$ time window $\times$ on-time rate $\times$ refundability --- and force a 2--3 plan comparison plus an executable order. \emph{Down}: fix destination / date, restrict to query + one booking.

\textbf{4. Concrete directions.} (A) \emph{Weather-triggered rebook}: check both candidate cities' weather, pick the safe one, place a refundable flight + ticket pair under a total-price cap. (B) \emph{Multi-leg + hotel}: Wuhan $\to$ Yinchuan / Zhongwei high-speed-rail or flight + 1 night hotel + next-day attraction with the earliest-arrival constraint and a total-budget cap.
\end{coachbox}

\begin{assistantbox}
\textbf{[1]} Following Coach direction~(A): I'll author a ``weather-triggered rebook'' task --- a Wuhan-departing traveller who must pick between Yinchuan and Ordos and book a flight + attraction ticket under a tight budget. Let me first probe the weather, attraction and flight catalogues for the dates in scope.
\end{assistantbox}

\begin{toolbox}
\textbf{[2]} \texttt{weather(city="Ordos", date\_start="2027-04-07", date\_end="2027-04-09")} $\Rightarrow$ \textbf{Apr.~8 \emph{sandstorm}}, 8--15\,$^\circ$C.\\
\texttt{weather(city="Yinchuan", ...)}\;$\Rightarrow$ Apr.~8 \emph{sunny}, 8--20\,$^\circ$C.\\
\texttt{weather(city="Zhongwei", ...)} $\Rightarrow$ Apr.~8 sunny, 10--22\,$^\circ$C.
\end{toolbox}

\begin{toolbox}
\textbf{[3]} \texttt{attractions\_search\_recommend(city="Ordos")} $\Rightarrow$ \emph{Xiangshawan} \texttt{A00005}; \emph{Kubuqi Desert} \texttt{A00007}.\\
\texttt{attractions\_search\_recommend(city="Zhongwei")} $\Rightarrow$ \emph{Shapotou National Nature Reserve} \texttt{A00006}.
\end{toolbox}

\begin{toolbox}
\textbf{[4]} \texttt{flights\_search(from="Wuhan", to="Yinchuan", date="2027-04-07")} $\Rightarrow$ \texttt{CA1834} 09:15--11:30, \texttt{HU7832} 14:25--16:40.\\
\texttt{flights\_search(from="Wuhan", to="Ordos", date="2027-04-07")} $\Rightarrow$ \texttt{MU2156} 08:30--10:45, \texttt{3U8756} 19:20--21:35.
\end{toolbox}

\begin{toolbox}
\textbf{[5]} \texttt{get\_ota\_attraction\_info / get\_ota\_flight\_info(...)} $\Rightarrow$ adult tickets \textyen 80--380; Wuhan$\to$Yinchuan economy fare \textyen 780; Wuhan$\to$Ordos economy fare \textyen 720--850.
\end{toolbox}

\begin{assistantbox}
\textbf{[6]} With both \emph{conflict constraints} (sandstorm vs sunny) and a \emph{tight budget} now grounded in real catalogue entries, the rubric-hit points are precise: the Solver must (i) explicitly consult the weather tool, (ii) deduce that Yinchuan is the safe destination, (iii) keep the total under \textyen 1500, and (iv) execute a flight order \emph{and} an attraction order. Emitting the final JSON.
\end{assistantbox}

\begin{instructionbox}
\textbf{[7] Generated task (emitted by the Generator):}\\[2pt]
\textit{Instructions:} ``I'm an engineer, I plan to depart Wuhan on 2027-04-07 for a Northwest trip. I'm undecided between Yinchuan and Ordos --- please check the weather on 2027-04-08 for both cities first. My departure date is 2027-04-07 (Mon), with 2 days of travel; flight + ticket combined budget must be $\le$\textyen 1500; flight should prefer high on-time rate, ticket should be value-for-money not the most expensive; finally place a flight order and an attraction-ticket order for me.''\\[3pt]
\textit{Rubrics (8):}
\begin{enumerate}
\setlength{\itemsep}{0pt}
  \item Agent queries the 2027-04-08 weather of both Yinchuan and Ordos.
  \item Agent recognises that Ordos has a sandstorm on Apr.~8 and recommends Yinchuan.
  \item Agent's chosen flight is Wuhan~$\to$~Yinchuan \texttt{CA1834}.
  \item Chosen departure date is 2027-04-07.
  \item Chosen seat class is economy (within budget).
  \item Ticket~+~flight total stays $\le$\textyen 1500.
  \item Agent creates the flight order via \texttt{create\_flight\_order}.
  \item Agent creates the attraction order via \texttt{create\_attraction\_order}.
\end{enumerate}
\end{instructionbox}

\noindent\textbf{Reward outcome.} The Generator's meta-reward (which captures \emph{format validity}, \emph{Solver-success-rate calibration around $K/2$}, and \emph{coverage of expected tools}) reaches its curriculum maximum on this rollout, signalling that the emitted task lies on the Solver's capability boundary --- non-trivial yet solvable when the weather check is performed correctly, and unsolvable when the Solver skips the weather query.

\subsection*{Trajectory F --- ECR-Travel Generator: Guangzhou-Shenzhen 2D1N}
\phantomsection
\label{traj:gen-travel}

\begin{coachbox}
\textbf{1. Failure diagnosis.} Past Generator outputs often demanded data that the toolkit cannot provide (real-time weather forecasts, actual ticket prices, exact opening hours), so the Solver could never satisfy the rubric. Cross-city multi-day queries also tended to be either all-pass or all-fail, with too many degrees of freedom.

\textbf{2. Diversity to explore.} Same-city multi-day (no inter-city leg), short regional trips ($\le$1--2~hr high-speed-rail), child / elder-friendly / accessibility, hard budget caps, earliest / latest-train constraints, rain-fallback itineraries.

\textbf{3. Difficulty knobs.} \emph{Down}: ask only for ``transport + POI order + lodging area'', use ``morning / afternoon'' instead of clock times. \emph{Up}: add transfer constraints (latest-arrival, earliest-departure), force two-plan comparison (cost-saving vs time-saving), require at least two distinct tool calls feeding into a comparison.

\textbf{4. Concrete directions.} (A) \emph{Guangzhou--Shenzhen, 2D1N}: compare two \texttt{train} candidates + verify the opening hours of two Shenzhen POIs. (B) \emph{Hangzhou, 3 days}: West Lake / Lingyin / Songcheng + rainy-day fallback (museum POI), use \texttt{poi\_search} only.
\end{coachbox}

\begin{assistantbox}
\textbf{[1]} Following Coach direction~(A): I'll author a Guangzhou-Shenzhen 2-day-1-night itinerary that requires the Solver to (i) recommend two specific train trips, (ii) verify the opening hours of two Shenzhen POIs, and (iii) check one driving distance --- all tool-grounded, all falsifiable.
\end{assistantbox}

\begin{toolbox}
\textbf{[2]} \texttt{poi\_search(address="Chimelong Tourist Resort, Guangzhou")} $\Rightarrow$ \texttt{(113.32, 23.00)}, the standard outbound anchor.\\
\texttt{poi\_search(address="Window of the World, Shenzhen")} $\Rightarrow$ \texttt{(113.98, 22.54)}, opens 15:00--21:30 on a typical winter weekday.\\
\texttt{poi\_search(address="Happy Valley, Shenzhen")} $\Rightarrow$ opens 09:00--22:00.
\end{toolbox}

\begin{toolbox}
\textbf{[3]} \texttt{search\_train\_tickets(date="2024-01-15", from\_city="Guangzhou", to\_city="Shenzhen")} $\Rightarrow$ multiple morning trains (departures 06:30--10:00, arrival $\le$2\,h) and evening returns peaking around 19:00--21:00; fares \textyen 60--150.
\end{toolbox}

\begin{toolbox}
\textbf{[4]} \texttt{direction(origin="Chimelong", destination="Happy Valley", mode="driving")} $\Rightarrow$ $\approx 114$~km, $\approx 1.5$\,h drive at typical traffic.
\end{toolbox}

\begin{assistantbox}
\textbf{[5]} All four tool returns are concrete and consistent: opening hours of both Shenzhen POIs are known, the train catalogue contains qualifying outbound \emph{and} return trips, and the driving distance can be cross-checked. The hard-constraint corner of the rubric (``return train must arrive in Guangzhou before 21:00'') is satisfiable but not trivial. Emitting the final JSON.
\end{assistantbox}

\begin{instructionbox}
\textbf{[6] Generated task (emitted by the Generator):}\\[2pt]
\textit{Query:} ``Planning a 2D1N trip from Guangzhou to Shenzhen on 2024-01-15: morning depart Chimelong Tourist Resort in Guangzhou, arrive Shenzhen by noon, afternoon visit Window of the World; day 2 morning visit Happy Valley, return to Guangzhou before 21:00 in the evening. Please recommend two appropriate train trips (one outbound, one return; return must arrive in Guangzhou before 21:00), verify the opening hours of Window of the World and Happy Valley fit the schedule, and report the driving distance from Chimelong to Happy Valley on day 1.''\\[3pt]
\textit{Expected tools:} \texttt{search\_train\_tickets} (Guangzhou $\to$ Shenzhen, 2024-01-15) \quad+\quad \texttt{poi\_search} (Window of the World) \quad+\quad \texttt{poi\_search} (Happy Valley) \quad+\quad \texttt{direction} (Chimelong $\to$ Happy Valley, driving).\\[3pt]
\textit{Rubrics (7):}
\begin{enumerate}
\setlength{\itemsep}{0pt}
  \item Recommended \emph{outbound} train departs no later than 08:00, so it arrives in Shenzhen before noon.
  \item Recommended \emph{return} train departs Shenzhen before 21:00 and arrives Guangzhou before 21:00.
  \item Return-train fare $\le$\textyen 150.
  \item The opening hours of Window of the World cover the afternoon window (15:00--21:30).
  \item The opening hours of Happy Valley cover the morning window (09:00--22:00).
  \item The driving distance from Chimelong to Happy Valley is reported as $\approx 114$~km.
  \item The proposed schedule is internally consistent (morning Chimelong $\to$ noon arrive Shenzhen $\to$ afternoon Window of the World $\to$ day-2 morning Happy Valley $\to$ evening return Guangzhou).
\end{enumerate}
\end{instructionbox}

\noindent\textbf{Reward outcome.}
The Generator attains the maximum reward: the generated task is a well-formed JSON object; all \texttt{expected\_tools} are invoked during task authoring with valid arguments; the rubric set places the task near the Solver's capability boundary, with the empirical Solver success count close to $K/2$ under the curriculum; and each rubric clause is grounded in, and falsifiable against, observable tool returns.

\end{document}